\documentclass[12pt]{article}

\usepackage{orcidlink,lmodern}
\usepackage{graphicx, float, caption}
\usepackage{natbib}
\usepackage{authblk} 
\usepackage{multirow}
\usepackage{hyperref}

\usepackage{amsmath,amssymb,amsfonts,amsthm}%
\usepackage{xcolor}%
\usepackage{manyfoot}%
\usepackage{algorithm2e}%
\usepackage{ulem}
\usepackage{mathtools} 
\usepackage{breqn} 
\usepackage{bbm} 
\usepackage{natbib}
\usepackage{mathrsfs}
\usepackage{array}
\usepackage{booktabs}
\usepackage{tabularx}
\usepackage{longtable}
\usepackage{url}
\usepackage{anyfontsize} 
\usepackage{silence} 
\usepackage{makecell}
\usepackage{algorithmic}

\newcommand{\bX}{\boldsymbol{X}}

\newcommand\maT{\mathcal{T}}
\newcommand\maE{\mathbb{E}}
\newcommand\maR{\mathbb{R}}
\newcommand\maI{\mathcal{I}}
\newcommand\maD{\mathcal{D}}
\newcommand\maS{\mathcal{S}}
\newcommand\con{\textnormal{\textsc{con}}}
\newcommand\lin{\textnormal{\textsc{lin}}}
\newcommand\pcon{\textnormal{\textsc{pcon}}} 
 
\newcommand\plin{\textnormal{\textsc{plin}}}

\newtheorem{theorem}{Theorem}
\newtheorem{lemma}{Lemma}

\newtheorem{remark}{Remark}%

\definecolor{orange1}{RGB}{255,128,0}
\definecolor{purple2}{RGB}{102,0,204}
\definecolor{blue}{RGB}{0,0,255}
\definecolor{red}{RGB}{255,0,0}

\begin{document}

\def\spacingset#1{\renewcommand{\baselinestretch}
{#1}\small\normalsize} \spacingset{1}

\title{A Powerful Random Forest Featuring Linear Extensions (RaFFLE)}

\author[1]{Jakob Raymaekers}
\author[2]{Peter J. Rousseeuw}
\author[1]{Thomas Servotte}
\author[1]{\\Tim Verdonck}
\author[1,2]{Ruicong Yao}

\affil[1]{Department of Mathematics, University 
          of Antwerp, Belgium}
\affil[2]{Section of Statistics and Data Science, 
          University of Leuven, Belgium}

\setcounter{Maxaffil}{0}
\renewcommand\Affilfont{\itshape\small}
\date{February 14, 2025}           
  \maketitle

\bigskip
\begin{abstract}
Random forests are widely used in regression. However, the decision trees used as base learners are poor approximators of linear relationships. To address this limitation we propose RaFFLE (Random Forest Featuring Linear Extensions), a novel framework that integrates the recently developed PILOT trees (Piecewise Linear Organic Trees) as base learners within a random forest ensemble. PILOT trees combine the computational efficiency of traditional decision trees with the flexibility of linear model trees. To ensure sufficient diversity of the individual trees, we introduce an adjustable regularization parameter and use node-level feature sampling. These modifications boost the accuracy of the forest. We establish theoretical guarantees for the consistency of RaFFLE under weak conditions, and its faster convergence when the data are generated by a linear model. Empirical evaluations on 136 regression datasets demonstrate that RaFFLE outperforms the classical CART and random forest methods, the regularized linear methods Lasso and Ridge, and the state-of-the-art XGBoost algorithm, across both linear and nonlinear datasets. By balancing predictive accuracy and computational efficiency, RaFFLE proves to be a versatile tool for tackling a wide variety of regression problems. 
\end{abstract}

\noindent {\it Keywords:} 
Algorithm; Consistency; Linear Model Trees; Machine Learning; Regression.

\section{Introduction}
Regression random forests are versatile ensemble learning techniques that have gained popularity due to their simplicity and strong predictive performance. 
At the core of a regression random forest, decision trees like CART \cite{cart} act as the base learners. Such a decision tree is a hierarchical method where data is split into subsets at each node, based on the value of a feature, and continuing until terminal nodes (leaves) are reached. Each leaf yields a prediction based on the average response of the training cases in that leaf, and the overall tree is thus a piecewise constant fit where the pieces are (hyper)rectangles in the original predictor space. However, a single decision tree can be sensitive to small changes in the training data, leading to high variance and overfitting.

Random forests, originally introduced in \cite{breiman2001random}, address this limitation by building a collection of decision trees, each trained on a different sample of the data, with the final prediction obtained by aggregating the predictions from all trees. This ensemble approach reduces variance, leading to more stable and reliable predictions, especially when dealing with noisy data or complex relationships.

Two key components make random forests effective: bootstrap aggregating (`bagging') and random feature selection. With bagging, each tree in the forest is trained on a bootstrap sample of the dataset, which is created by random sampling with replacement. This technique introduces diversity among the trees, as each tree receives a slightly different version of the data. Random feature selection means that during the training of each decision tree, only a random subset of features is considered for splitting at each node. This decorrelates the trees, preventing any single feature from dominating the decision-making process and ensuring that the ensemble captures a broad spectrum of data characteristics.

Random forests are particularly well-suited for regression tasks due to their ability to fit complex, non-linear relationships between input variables and the target variable. They can handle high-dimensional data, mixed data types (numerical and categorical), missing values, and interactions between features without explicit specification. Additionally, random forests inherently provide measures of feature importance, giving insights into the factors driving the predictions. While there have been many advances in regression such as Gradient Boosted Trees \citep{xgboost, lightgbm} and Neural Networks \citep{tabnet}, random forests remain popular and are often found to outperform more complex methods over a wide range of tasks \citep{rf_performance_rodriguez2015, FERNANDEZDELGADO201911, zamo2014benchmark}.

Despite their advantages, traditional random forests have limitations, particularly in handling datasets with linear characteristics. This limitation has driven research into new types of decision trees and ensembles that can extend or replace the standard random forest framework, addressing specific challenges or enhancing predictive performance in particular contexts. An overview of ensemble approaches for regression can be found in \cite{mendes2012ensemble}. An ensemble of generalized linear models (GLMs) was proposed by \cite{song2013random}, who found that it performs similarly to a random forest. Ensembles of regression trees with some linear characteristics, also called linear model trees, were investigated by \cite{ao2019linear}, \cite{rodriguez2010experimental} and \cite{li2011learning}, illustrating the potential of random forests composed of more expressive base learners. \citet{freund1996experiments} proposed boosting as a versatile framework, laying the groundwork for Gradient Boosted Decision Trees (GBDT) such as XGBoost \citep{xgboost} and LightGBM \citep{lightgbm}. Later, \citet{shi2019gradient} proposed a GBDT method using piecewise linear model trees as base learners, highlighting the effectiveness of combining boosting with alternative base learners.

In this paper we propose RaFFLE (\textbf{Ra}ndom \textbf{F}orest \textbf{F}eaturing \textbf{L}inear \textbf{E}xtensions), a random forest of linear model trees. The latter are of the PILOT type, which stands for \textbf{PI}ecewise \textbf{L}inear \textbf{O}rganic \textbf{T}ree \cite{raymaekers2024pilot}.  PILOT trees combine the computational efficiency and flexibility of classical decision trees with the expressiveness of linear model trees, making them particularly well-suited to handle datasets where linear relationships play a substantial role. By leveraging the strengths of PILOT, RaFFLE offers several advantages over traditional random forests and other ensemble methods. These include improved handling of both linear and nonlinear data structures and a balance between predictive accuracy and computational efficiency. We develop RaFFLE as a solid framework for regression tasks, and rigorously evaluate its theoretical properties and empirical performance.

The paper is structured as follows. In Section \ref{pilot_sec:methodology} we describe the PILOT algorithm and its properties. We also explain the modifications adapting PILOT for use in a random forest, including the introduction of a regularization parameter, node-level feature sampling, and computational speedups.
 
In Section \ref{pilot_sec:theory} we provide a rigorous theoretical analysis of RaFFLE. We prove its consistency under weak conditions as well as its convergence rate on linear data, and discuss the implications for predictive performance. We also derive its computational complexity.
 
In Section \ref{pilot_sec:results} we present an empirical evaluation of RaFFLE on 136 regression datasets, comparing its performance with that of CART and the traditional random forest, the state-of-the-art XGBoost algorithm, and the Lasso and Ridge regularized linear methods. The results showcase RaFFLE's superior accuracy across datasets with diverse characteristics, highlighting its versatility.
 
Finally, in Section \ref{pilot_sec:conclusion} we summarize the key contributions of RaFFLE and propose directions for future work.

\section{Methodology}\label{pilot_sec:methodology}
We first provide an overview of the PILOT algorithm, and then describe the modifications to make PILOT a suitable base learner in a random forest ensemble.

\subsection{Linear model trees by PILOT}

The popular piecewise constant binary trees require little computational cost, but they yield inefficient approximations of smooth continuous functions. To resolve this issue, linear model trees have been proposed \citep{torgo1997functional, loh2011classification, stulp2015many}. They preserve the tree-like structure for partitioning the predictor space, but in the leaf notes they allow for linear combinations of the variables. These model trees offer greater flexibility while preserving the simple structure of traditional regression trees. Until recently, one of the main drawbacks of linear model trees was their high computational cost. As a result, they were rarely used to analyze large datasets, and  incorporating them into ensemble methods would lead to infeasible computational costs.

The PILOT (\textbf{PI}ecewise \textbf{L}inear \textbf{O}rganic \textbf{T}ree) method \cite{raymaekers2024pilot} is a novel linear model tree algorithm designed to address the limitations of earlier approaches for fitting linear model trees. Most notably, PILOT runs with the same time complexity as the classical CART algorithm for binary decision trees. This gives PILOT the ability to fit very large datasets and opens the door to the integration of linear model trees in ensemble methods such as random forests. In addition to a relatively low computational cost, PILOT improves upon existing algorithms for linear model trees by efficiently guarding against extrapolation, and by incorporating on-the-fly regularization to avoid overfitting. This regularization makes it unnecessary to prune the tree afterward, hence the adjective `organic'. The method enjoys strong theoretical support, as it is consistent on additive data and achieves faster convergence rates when the data are generated by a linear model. Empirically, PILOT outperforms CART when the underlying relationship between predictors and response is roughly continuous and smooth, while performing on par otherwise. 

\begin{figure}[!ht]
    \centering
    \begin{tabular}{ccccc}
        \textbf{p}iecewise & & & \textbf{b}roken & 
        \textbf{p}iecewise \\ 
        \textbf{con}stant & \textbf{con}stant & 
        \textbf{lin}ear & \textbf{lin}e &
        \textbf{lin}ear \\   
        \includegraphics[width = 0.17\textwidth]{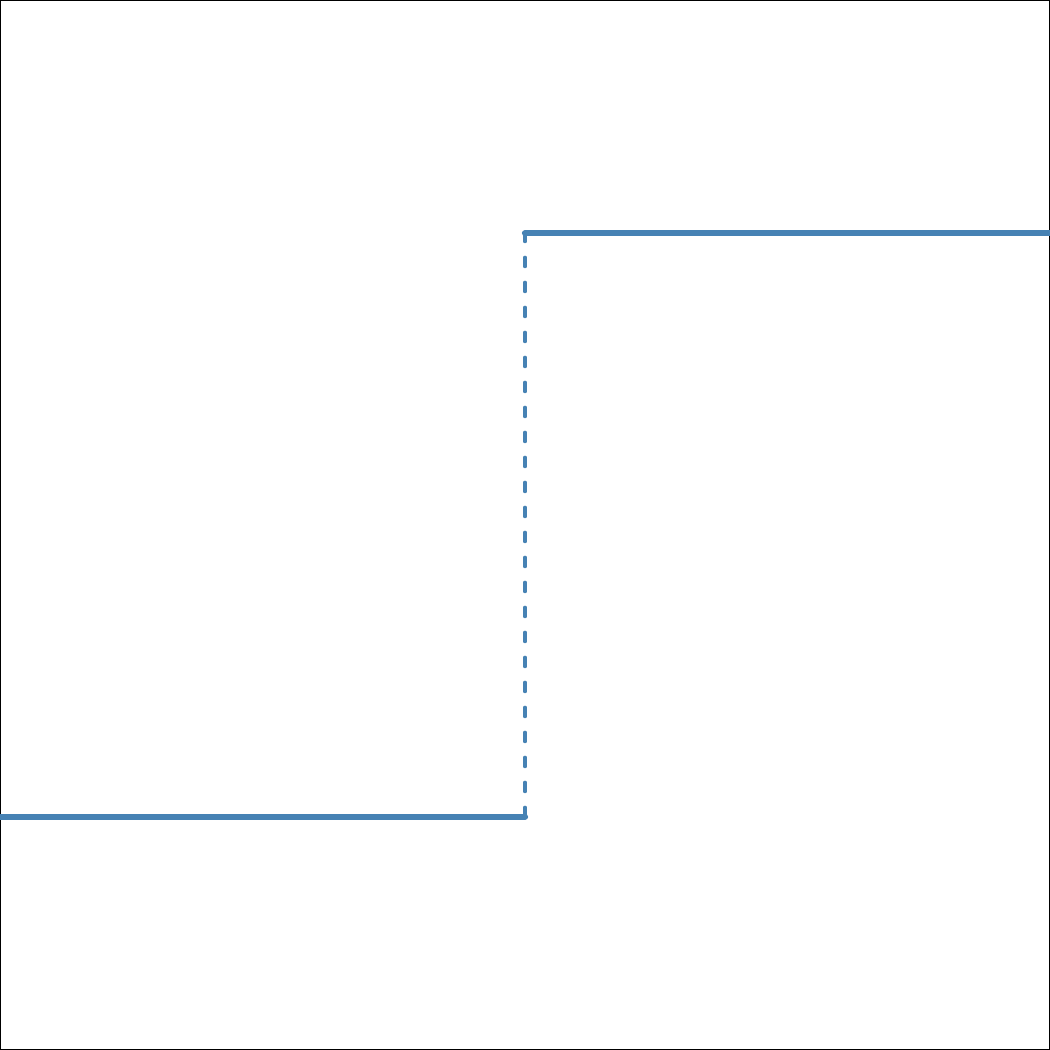} &
        \includegraphics[width = 0.17\textwidth]{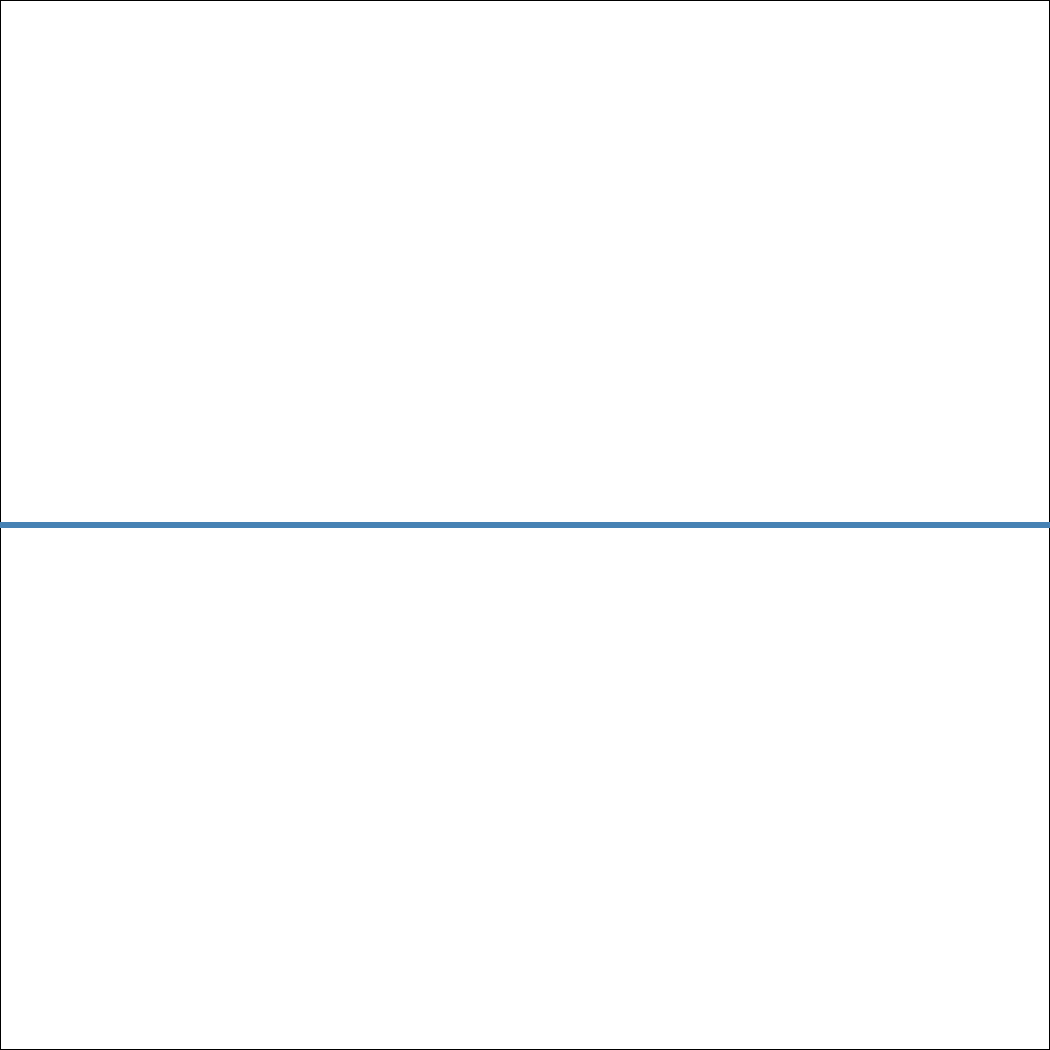} &
        \includegraphics[width = 0.17\textwidth]{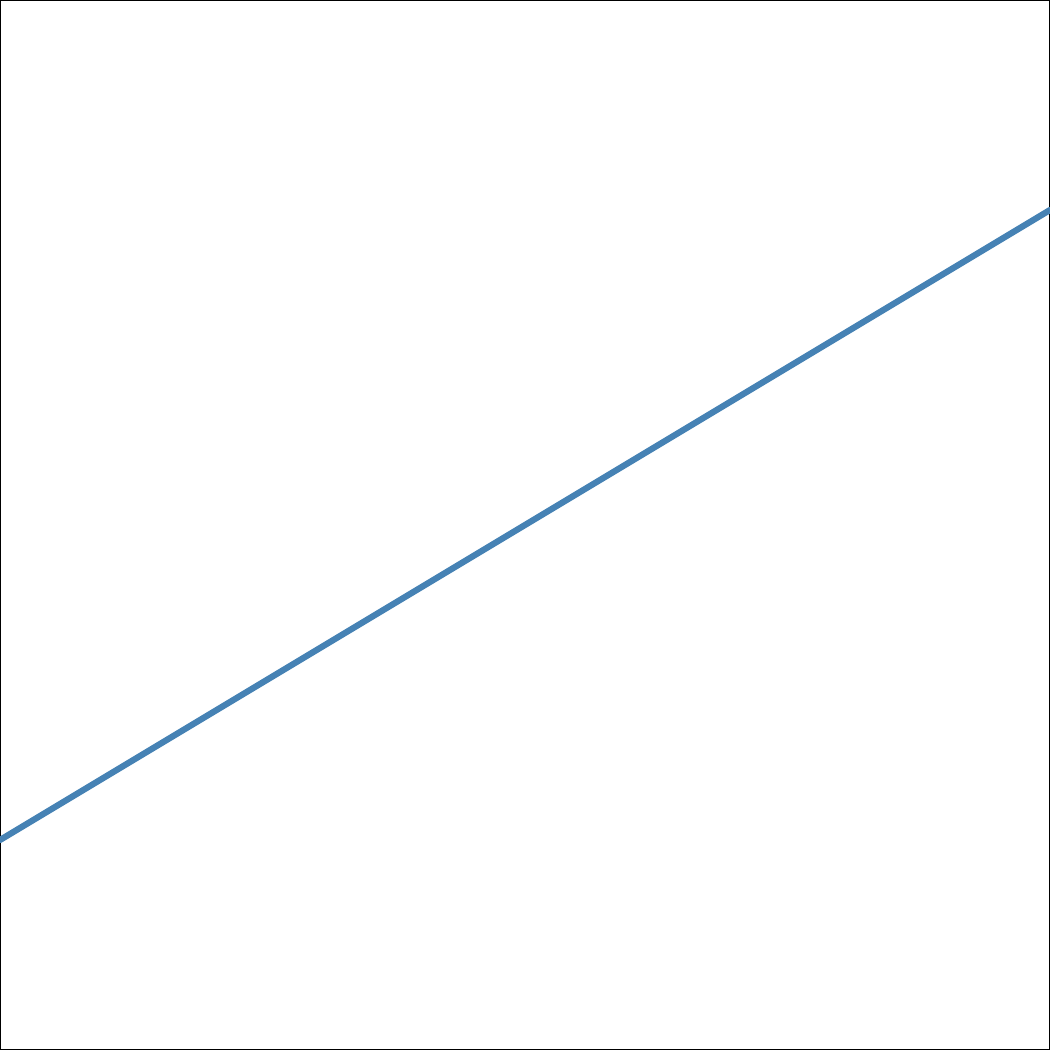} &
        \includegraphics[width = 0.17\textwidth]{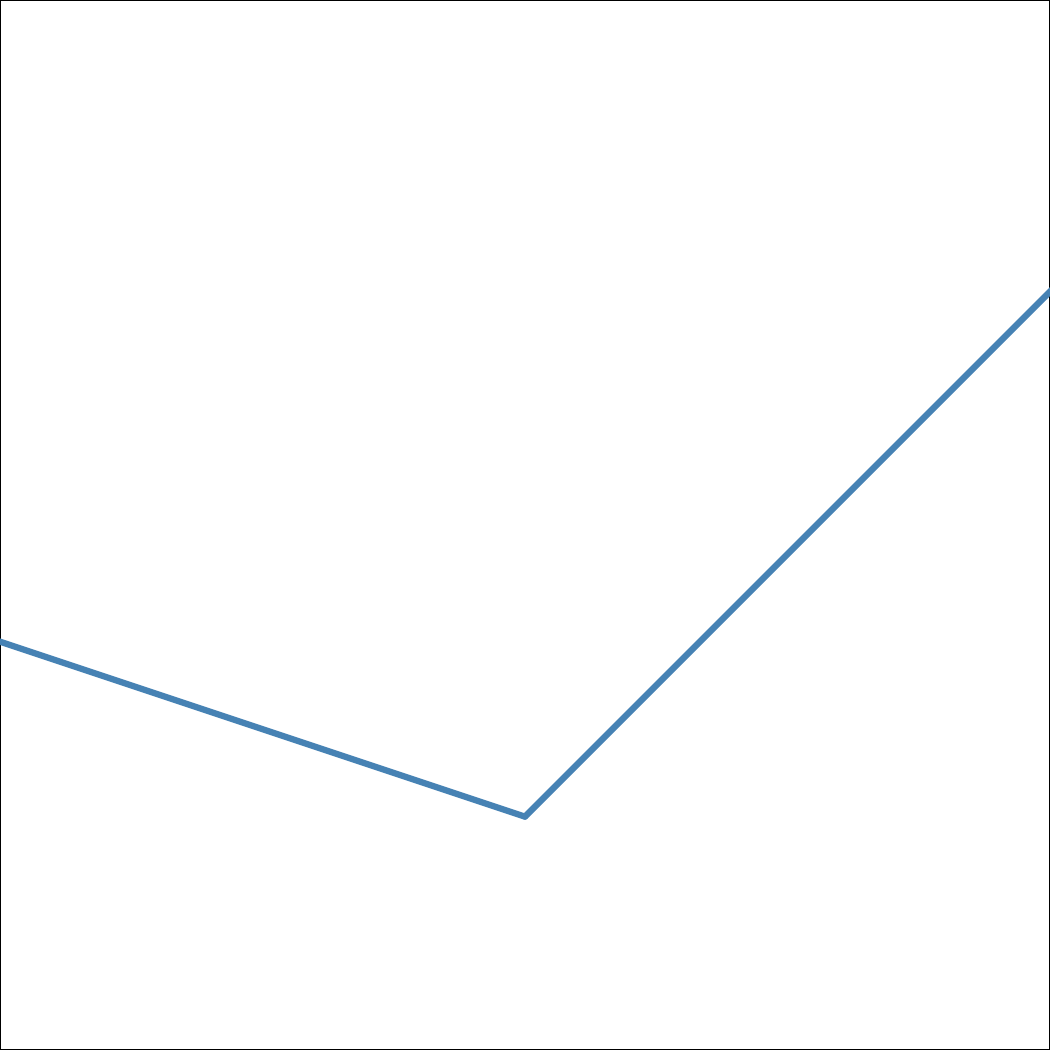} &
        \includegraphics[width = 0.17\textwidth]{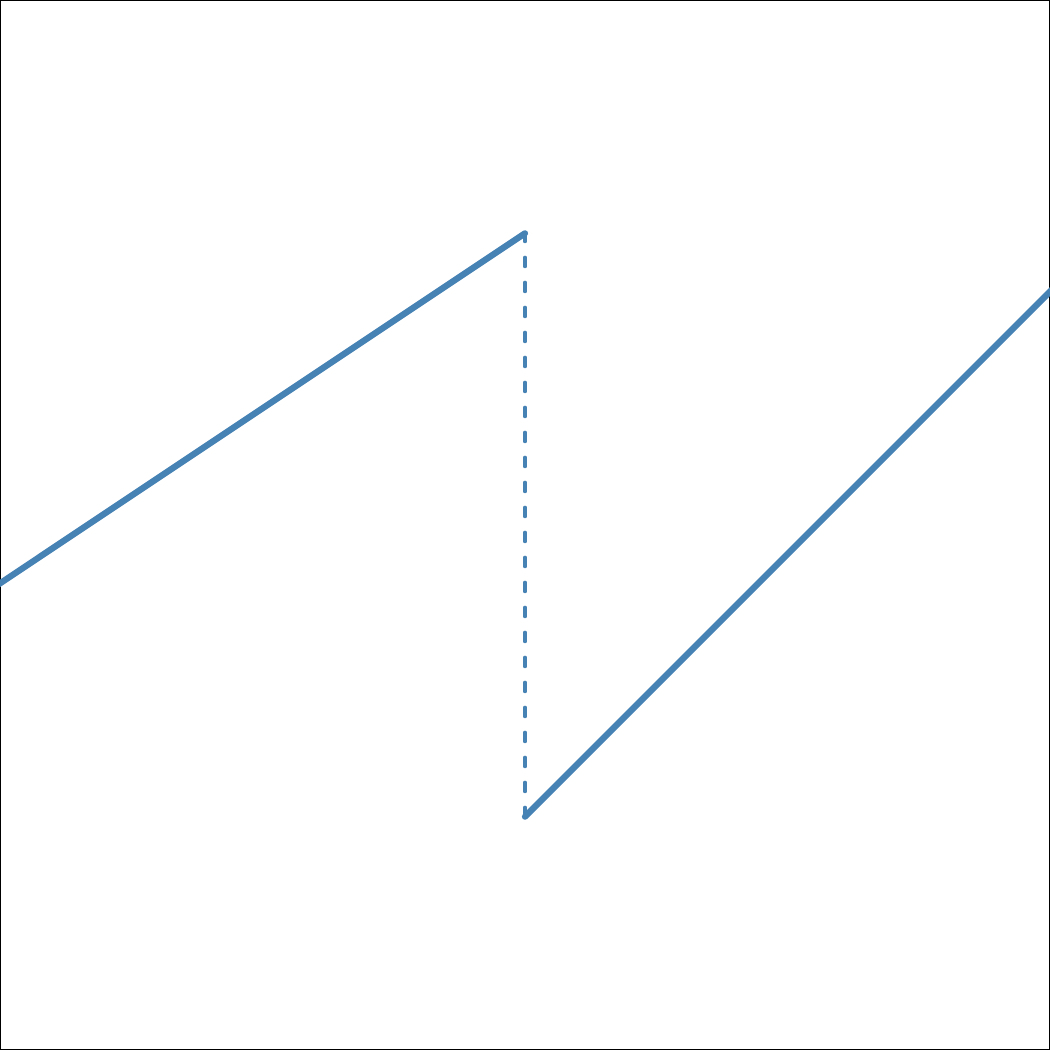}\\
    \end{tabular}
\caption{The five regression models used in PILOT:
  \textsc{pcon}, \textsc{con}, \textsc{lin}, 
  \textsc{blin} and \textsc{plin}.} 
\label{fig:fivemodels}
\end{figure}

The PILOT algorithm takes a very similar approach to CART in that it builds the regression tree through a greedy top-down process. More precisely, PILOT starts with all observations in one node, and splits this node if that improves the fit. When split, the observations within a node are divided between the two child nodes, and the process is repeated in each of the child nodes separately. Unlike CART however, PILOT incorporates linear models in this process, in the following way. Instead of only considering a piecewise constant model on each predictor, PILOT considers five different models and selects the best one. 
These five models are shown in Figure~\ref{fig:fivemodels}.
The piecewise constant model (\textsc{pcon}) is that of CART, and it also applies to categorical regressors. A node that isn't split has a constant fit (\textsc{con}). To these models PILOT adds three models with linearity. The first type is the univariate linear model (\textsc{lin}), which does not split the node. Another is the `broken line' (\textsc{blin}) which is continuous, and the most general is the piecewise linear model (\textsc{plin}). 

\begin{figure}[!ht]
\center{\includegraphics[scale=0.8]{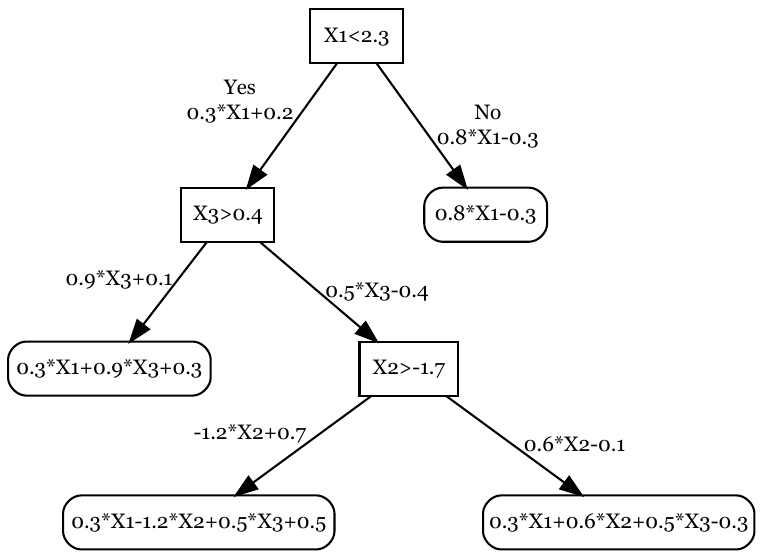}}
\caption{An example of a PILOT tree} 
\label{fig:treeplot}
\end{figure}

After a model is fitted, the residuals of that fit are passed on as the new response for the next step. When  \textsc{con} is fit, the recursion in that node stops. Only \textsc{pcon}, \textsc{blin}, and \textsc{plin} split the node into two child nodes. When \textsc{lin} is fit, the node is not split but the residuals are updated and passed on to the next step. For a given feature, all five models can be evaluated in one pass through the feature, just as in CART. The coefficients of the linear model in a leaf node are obtained by adding the coefficients encountered in the path from root to leaf node, as illustrated in Figure~\ref{fig:treeplot}.

Of course, as the \textsc{plin} model is strictly more general than the other models, selecting the model type solely based on the largest reduction in residual sum of squares would lead to only selecting \textsc{plin} models. To avoid this, in each node PILOT uses a Bayesian Information Criterion (BIC) to balance model complexity with goodness of fit. It is given by
\begin{equation} \label{eq:BIC}
  \mbox{BIC} = n\log\Big(\frac{\mathrm{RSS}}{n}\Big)+\nu\log(n)
\end{equation}
which depends on the residual sum of squares $\mathrm{RSS}$, the number of cases $n$ in the node, and the degrees of freedom $\nu$ of the model. PILOT uses $\nu = 1, 2, 5, 5, 7$ for the \textsc{con}, \textsc{lin}, \textsc{pcon}, \textsc{blin} and \textsc{plin} models.
When fitting a node, PILOT chooses the combination of a feature and a regression model that yields the lowest BIC value. This avoids overfitting in a computationally efficient way, in contrast to approaches that require costly post-hoc pruning.\\

\subsection{PILOT in a random forest}

While PILOT did very well in empirical studies, its performance as a standalone method does not automatically make it a good base learner for a random forest. In particular, PILOT was designed with a lot of emphasis on avoiding overfitting. This was needed in view of the large flexibility of combining recursive partitioning and linear model fits. Due to this implicit regularization, a PILOT tree typically has a smaller variance than a CART tree trained with the same maximum depth. Random forests leverage base learners with relatively large variance, so the out-of-the-box PILOT is less suited as a base learner.

In order to make PILOT suited as a base learner in a random forest, we thus need to curtail its inherent regularization and allow the linear model trees to have more variability. To achieve this, we propose two major changes. The first is to adjust the BIC criterion used to select the best model fit in a given node, which balances goodness of fit and model complexity. For this we propose to replace the BIC criterion of equation~\eqref{eq:BIC} by
\begin{equation} \label{eq:BICalpha}
  \mbox{BIC}_\alpha = 
  n\log\Big(\frac{\mathrm{RSS}}{n}\Big) +
  \nu_\alpha \log(n)
\end{equation}

\vspace{1mm}
\noindent where $\nu_\alpha \coloneqq 1 + \alpha(\nu - 1)$ and $0 \leqslant \alpha \leqslant 1$ is a tuning parameter. So when $\alpha = 1$, the default $\nu$ values from the standalone PILOT are used. For $\alpha=0$ all model types have equal $\nu$ values of 1, and we would only fit \textsc{plin} models. The lower the value of $\alpha$, the more the tree will tend to overfit the data, which is desired to some extent as we want to increase the variability of the base learner in the random forest.

The second major change to the original PILOT algorithm is that we disable the broken line (\textsc{blin}) model. The broken line is a regularized version of the piecewise linear (\textsc{plin}) model. Removing it typically results in proportionally more \textsc{plin} fits, increasing the variance of the tree. As an additional advantage, the \textsc{blin} model is the most computationally demanding one.  Therefore removing this model results in a significant speed-up during training, which turned out not to lose performance in a random forest. We emphasize that when fitting a standalone PILOT tree, it remains advised to keep \textsc{blin} in the set of models.

In addition to partially curtailing the regularization, we made several small adjustments. In particular, after shifting the response up or down to give it a symmetric range $[-B,B]$, we truncate all predictions to the interval $[-B, B]$ instead of $[-1.5B, 1.5B]$ as originally proposed. This means that the individual trees cannot generate predictions outside of the range of the response in the training data. This option is safer and was found to perform the best in general. 

Overall, our modifications enhance the diversity of the PILOT trees in the random forest, reduce the computational cost of the trees, and safeguard against extrapolation. We now construct RaFFLE as a random forest of linear model trees trained by the modified PILOT algorithm. The two main building blocks of random forests are used. The first is that each tree in the ensemble is trained on a newly generated bootstrap sample of the data. The second is random feature selection. In the classical random forest, each tree is trained on a newly generated random subset of $q$ features where $q \leqslant p$. We go one step further, by newly generating a random subset of $q$ features at every node of every tree. The PILOT trees trained with both bootstrapping the data and this node-level random feature selection we call  \textit{random PILOT trees}, to distinguish them from the deterministic standalone PILOT trees. The pseudocode of RaFFLE is shown in Algorithm~\ref{algo:RaFFLE}.

\begin{algorithm}
\caption{The RaFFLE fitting algorithm} \label{algo:RaFFLE}
\begin{algorithmic}[1]
\REQUIRE \
\vspace{-2mm}
\begin{itemize}
\itemsep -3pt
    \item $X$ (Feature data)
    \item $y$ (Target values)
    \item \textsc{n\_estimators} (Number of PILOT trees)
    \item \textsc{n\_features\_tree} (Number of features used by each tree)
    \item \textsc{n\_features\_node} (Number of features used in the nodes of each tree)
    \item \textsc{alpha} (Tuning parameter for degrees of freedom of each node type)
    \item \textsc{max\_depth} (Maximum depth of trees, excluding \textsc{LIN} nodes)
    \item \textsc{max\_model\_depth} (Maximum depth of trees, including \textsc{LIN} nodes)
    \item \textsc{min\_sample\_fit} (Minimum number of cases required to fit any model)
    \item \textsc{min\_sample\_alpha} (Minimum number of cases required to fit a
    
    piecewise model)
    \item \textsc{min\_sample\_leaf} (Minimum number of cases required in leaf nodes)
\end{itemize}
\STATE $\textsc{n\_features\_node} \leftarrow \min(\textsc{\_features\_tree}, \textsc{n\_features\_node})$
\STATE Initialize $estimators$ as a list of
\textsc{n\_estimators} random PILOT trees that

use this parameter \textsc{n\_features\_node} 
\FOR{$i = 1$ to $n\_estimators$}
    \STATE Take a random bootstrap sample with row indices $r_i$
    \STATE Fit $estimators[i]$ on $(X[r_i], y[r_i])$
    using the above parameters
\ENDFOR
\end{algorithmic}
\end{algorithm}

We have implemented PILOT in \texttt{C++} with a \texttt{Python} interface using \texttt{pybind11}. RaFFLE is built in \texttt{Python} around this interface.

\section{Theoretical results}\label{pilot_sec:theory}

We now focus on the theoretical analysis of RaFFLE, by exploring its consistency and convergence rate.
 
The notations are aligned with \cite{raymaekers2024pilot}. In particular, $||x||_n$ and $\langle x,y\rangle_n$ denote the empirical squared norm and the empirical inner product for $n$-dimensional vectors $x$ and $y$. 
The set $\maT_k^{(m)}$ consists of the tree nodes at depth $k$ plus the leaf nodes of depth lower than $k$, in the $m$-th random PILOT tree of the RaFFLE method $\mathcal{R}_{M,K} = (\maT^{(1)}_{K},\dots,\maT^{(M)}_{K})$ with $M$ trees and maximum depth $K$. 
The prediction of the PILOT tree at depth $k$ is denoted as $\hat f(\maT_k^{(m)})$, and $\hat f(\mathcal R_{M,K})$ is the prediction of RaFFLE. 
The training error at depth $k$ is denoted as $R_k^{(m)}:=||Y-\hat f(\maT_k^{(m)})||^2_n - ||Y-f||^2_n$ where $f$ is the underlying function and $Y$ is the response variable. We occasionally omit $m$ for notational simplicity.

We can rewrite $R_k$ as $\sum_{T\in \mathcal{T}_K}w(T)R_k(T)$ with weights $w(T):=t/n$ where $t$ is the number of cases in node $T$, and $R_k(T):=||Y-\hat f(\mathcal{T}_k)||^2_t-||Y-f||^2_t$. Then we immediately have $R_{k}=R_{k-1}-\sum_{T\in \mathcal{T}_{k-1}}w(T)\Delta^{k}(T)$ where
$$\Delta^{k}(T):=||Y-\hat{f}(\mathcal{T}_{k-1})||^2_t\;-\;t_l||Y-\hat{f}(\mathcal{T}_{k})||^2_{t_l}/t\;-\;t_r||Y-\hat{f}(\mathcal{T}_{k})||^2_{t_r}/t$$
is the impurity gain on $T$, and $t_l$ and $t_r$ denote the number of cases in the left and right child of $T$. In particular, if the $\lin$ model is selected, then by convention we let $t_r=0$ since there is no split. The design matrix of all $n$ cases is denoted as $\bX\in\maR^{n\times p}$. Given some tree node $T$ with $t$ cases, we denote by $\bX_T:=(X_{T_1},\dots,X_{T_t})^\top\in\maR^{t\times p}$ the restricted data matrix and by $X^{(j)}_T$ its $j$-th column. The variance of the $j$-th column is given by $\hat\sigma_{j,T}^2:=\sum_{k=1}^t(X_{T_k}^{(j)}-\overline{X^{(j)}_T})^2/t$. We also let $(\hat\sigma_{j,T}^u)^2$ be the classical unbiased variance estimates with denominator $t-1$. The $T$ is omitted when $T$ is the root node, or if there is no ambiguity. The total variation of a function $f$ on $T$ is denoted as $||f||_{TV(T)}$. For nonzero values $A_n$ and $B_n$ which depend on $n\rightarrow\infty$, we write $A_n\precsim B_n$ if and only if $A_n/B_n\le \mathcal{O}(1)$, and $\succsim$ is analogous. We write $A_n\asymp B_n$ if $A_n/B_n= \mathcal{O}(1)$. The complement of a set $A$ is denoted by $A^c$, and $\# A$ stands for the cardinality of $A$.

RaFFLE trains the $m$-th tree in the forest $\mathcal{R}_{M,K}$ on a bootstrapped sample of data points 
$\maD_{\maI_m}\subset \maD_n$ with indices $\maI_m$. In addition to that, at each node it uses only $q \leqslant p$ predictors that are selected at random. In the following, we use $\Xi_k$ to denote the random variable that selects the sets of predictors $\maS_{T}\subset\{1,\dots,p\}$, in every node $T$ at depth $k$, where $\#\maS_{T} = q$. Note that $\Xi_k$ is independent of the cases and the response but depends on $m$, but for notational simplicity we omit $m$ here. 

\subsection{Consistency}\label{subsec:consistency}
We assume the underlying function $f\in\mathcal{F}\subset L^2([0,1]^p)$ admits an additive form
\begin{equation} \label{eq:additive}
  f(X):=f_1(X^{(1)})+\dots+f_p(X^{(p)})
\end{equation}
where $f_j$ has bounded variation and $X^{(j)}$ is the $j$-th predictor. We define the total variation norm $||f||_{TV}$ of $f\in\mathcal{F}$ as the infimum of $\sum_{j=1}^p||f_j||_{TV}$ over all possible representations of $f$, and assume that the representation in \eqref{eq:additive} attains this infimum.
The consistency of a single PILOT tree was established in \citep{raymaekers2024pilot}.
Following \cite{klusowski2024large} we now extend this consistency result to the forest ensemble. 

We first derive the expected impurity gain of a random PILOT tree at each depth $k$ on the dataset $\maD_{\maI_m}$, and then aggregate the gains over $K$ steps for an error estimate of the entire tree. In the end we will take the expectation over all trees to prove the consistency of RaFFLE.

For the expected gain at depth $k$, it suffices to consider the conditional expectation on $\Xi_k|\Xi_{k-1}$ which captures the randomness of $\maS_T$ for any $T$ at depth $k-1$, i.e.\ $ \maE_{\Xi_k|\Xi_{k-1}}\left[\max_{j\in\maS_T}\widehat\Delta^{k}(\hat s_j,j,T)\right]$. Here, $\hat s_j$ is the optimal split point for feature $j$ in terms of the gain $\widehat\Delta^{k}$ in node $T$. We start with the situation when $\widehat\Delta^{k}$ stems from a  \textnormal{\textsc{pcon}} node and later generalize the results to arbitrary nodes. Lemma \ref{lem:pcon_step_est} gives a lower bound on the expected impurity gain of \textsc{pcon} for a specific node $T$ at depth $k$.

\begin{lemma}\label{lem:pcon_step_est}
Assuming the response variable is bounded in $[-B,B]$ and $R_{k-1}(T)>0$ in some node $T$, then the expected impurity gain of \textnormal{\textsc{pcon}} on this node satisfies
\begin{equation*}
   \maE_{\Xi_k|\Xi_{k-1}}\left[\max_{j\in\maS_T}\widehat\Delta^{k}_{\mbox{\tiny PCON}}(\hat s_j,j,T)\right]\geqslant \frac{qR_{k-1}^2(T)}{p(||f||_{TV}+2B)^2}
\end{equation*}
where $\hat s_j$ is the optimal splitting point of a \textsc{pcon} model fit to the observations in node $T$ using the $j$-th feature.
\end{lemma}

All proofs of this subsection are in Appendix A.

We next derive Lemma \ref{lem:bic} which leverages the BIC model selection to quantify the expected impurity gain of any selected model relative to that of \textsc{pcon}.

\begin{lemma}\label{lem:bic}
Consider some node $T$ with $t$ cases. For any random set of predictors $\maS_T$ induced by $\Xi_k|\Xi_{k-1}$ we let $\Delta_1$, $\Delta_2$ and $\nu_1$, $\nu_2$ be the impurity gains and degrees of freedom of two regression models on $T$ based on $\maS_T$. Let $R_T$ be the initial residual sum of squares in $T$. We have two results.
\begin{itemize}
\item Let $E_1$ be the event under $\Xi_{k}|\Xi_{k-1}$ that model 1 does better than \textnormal{\textsc{con}}, i.e.\ $BIC_{\mbox{\tiny{CON}}}> BIC_1$. Then we have
$t\maE[\widehat\Delta_1|E_1]/R_T> C(\nu_1,t)>0$ for some positive function $C$ depending on $\nu_1$ and $t$. Similarly $t\maE[\widehat\Delta_1|E_1^c]/R_T\leqslant C(\nu_1,t)$. Here, 
$$C(v,t) = \frac{1}{t}\Big(1-\exp\Big((\nu_{\mbox{\tiny CON}}-\nu_1)\frac{\log t}{t}\Big)\Big).$$
\item 
Let $E_2$ be the event that the selected model is not $\con$. Let $\widehat\Delta$ be the impurity gain of the selected non-$\con$ model based on the BIC criterion and the random predictors $\maS_T$ induced by $\Xi_{k}|\Xi_{k-1}$. Then we have
$\maE[\widehat\Delta|E_2] \geqslant \frac{1}{4}\maE[\Delta_{\pcon}|E_2]$.
\end{itemize}
\end{lemma}

\begin{remark}\label{rem:connode}
Note that in a random tree it is not necessarily true that after selecting \textnormal{\textsc{con}} at a node $T$, subsequent steps will also choose \textnormal{\textsc{con}}. This is because the random set $\maS_T$ may not include the optimal predictor $\widehat j_p$ that has an impurity gain using a non-$\con$ model. When the depth increases, $\widehat j_p$ and the corresponding model still have a chance to be selected. In our analysis we will therefore use Lemmas \ref{lem:pcon_step_est} and \ref{lem:bic} to estimate the impurity gain for nodes that include the optimal predictor $\widehat j_p$, otherwise, we lower bound the gain by zero. 
\end{remark}

We now obtain the error estimation of a single random PILOT tree in Theorem \ref{theorem:errorRandomPILOT}.

\begin{theorem}\label{theorem:errorRandomPILOT}
    Let $f\in \mathcal{F}$ with finite\ $||f||_{TV}$ and denote by $\hat f(\mathcal{T}_K)$ the prediction of a $K$-depth random PILOT tree. Suppose that $X\sim P$ on $[0,1]^p$ and the response variable is a.s.\ bounded in $[-B,B]$. Let $q$ denote the number of random predictors out of $p$. Then
\begin{equation}
\maE_{\Xi_k}(||Y-\hat f(\mathcal{T}_K)||^2_n)\, \leqslant ||Y-f||^2_n \, + \frac{4p(||f||_{TV}+2B)^2}{q(K+3)}+\maE_{\Xi_k}(R_{C_K^+})\;.
\end{equation}
Moreover, if we let $K=\log_2(n)/r$ with $r>1$, then $\maE_{\Xi_k}(R_{C_K^+})$ goes to zero at the rate $\mathcal{O}\left(\sqrt{\log (n)/n^{(r-1)/r}}\right)$.
\end{theorem}

Finally, we obtain the universal consistency of RaFFLE on additive data:
\begin{theorem}\label{theorem:consistency}
   Let $f\in \mathcal{F}$ with finite\ $||f||_{TV}$ and denote by $\hat f(\mathcal {R}_{M,K_n})$ the prediction of RaFFLE with M trees and depth $K_n$. Suppose $X\sim P$ on $[0,1]^{p_n}$ and the response variable is a.s.\ bounded. If the depth $K_n$ of each tree, the number of random predictors $q_n$ and the number of total predictors $p_n$ satisfy $q_nK_n/p_n\rightarrow\infty$ and $2^{K_n}p_n\log(np_n)/n\rightarrow0$, then RaFFLE is consistent, that is
\begin{equation}
\lim_{n\rightarrow\infty} \maE[||f-\hat f(\mathcal {R}_{M,K_n})||^2]=0.
\end{equation} 
\end{theorem}

\subsection{Convergence rate on linear data}\label{subsec:convergencelinear}

In this subsection we study the convergence rate of RaFFLE on linear data. As RaFFLE incorporates linear model trees, we expect the convergence rate to be better than the universal rate obtained in the previous section. The key to the proof is to tackle the randomness introduced by $\Xi_k$ and $\maD_{\maI_m}$. We assume Conditions 1 and 2 below.
\begin{itemize}
\item \textbf{Condition 1:} The PILOT algorithm stops splitting a node whenever
\begin{itemize}
\vspace{-0.2cm}
\item the number of cases in the node is less than $n_{\mbox{\tiny min}}=n^\delta$ for some $0<\delta<1$; 
\item the variance of some predictor is less than $2\sigma_0^2$ where $0<\sigma_0<1$; 
\item the volume of the hyperrectangle of the node is less than a threshold $\eta > 0$.
\end{itemize}
\item \textbf{Condition 2:} We assume that $X\sim P$ on $[0,1]^p$ and the error $\epsilon$ has a finite fourth moment. Moreover, for any hyperrectangle $C$ with volume up to $\eta$ we assume that\linebreak $\lambda_{min}(Cor(X|X\in C)) \geqslant 2\lambda_0>0$, where $Cor(X)$ is the correlation matrix.
\end{itemize}

We denote the least squares loss on the node $T$ by $L^*_n(T):=\min_{\hat\beta}||Y-\boldsymbol X_T\hat\beta||^2_T$ and the least squares loss on the full data by $L^*_n:=\min_{\hat\beta}||Y-\boldsymbol X\hat\beta||^2_n$. We further denote by $L^k_n:=||Y-\hat f(\maT_k)||^2_n$ the loss of a $k$-depth PILOT tree, and by $L^k_n(T):=||Y_T-\boldsymbol X_T\beta_T||^2_t$ its loss in node $T$ for some $\beta_T$ (we can write the loss like this because the prediction function is linear on $T$). When the depth $k$ is not important, we omit the superscript. We use the notation  $\boldsymbol{\tilde X}_T$ for the standardized predictor matrix obtained by dividing each (non-intercept) column $j$ in $\bX$ by $\sqrt{n}\hat\sigma^u_j$ where $(\hat\sigma^u_j)^2$ is the unbiased estimate of the variance of column $j$. 

We first derive the expected impurity gain of the $\lin$ model when $q$ predictors are selected randomly out of $p$.
\begin{lemma}\label{lem:lin_step_est}
Let $T$ be a node with $t$ cases with training loss $L^{(k-1)}_n(T)$, and $L^*_n(T)$ the training loss of least squares regression. In addition to Conditions 1 and 2 we assume that the residuals of the response on T have zero mean. 
The features in the node form $\boldsymbol X_T\in\mathbb R^{t\times p}$ which is a subset of $\boldsymbol X\in\mathbb R^{n\times p}$ randomly depending on $\boldsymbol X$, $\Xi_{k-1}$ and the noise. 
If the \textit{random} PILOT tree considers $q\leqslant p$ variables selected by $\Xi_k|\Xi_{k-1}$, then there exists a constant $C_{\lambda_0,\sigma_0,p}$ such that 
 $$P_{\boldsymbol X_T}\Bigg[\maE_{\Xi_k|\Xi_{k-1}}\left(\Delta_{\lin}\right)\geqslant \frac{q^2\lambda_0(L_n^{k-1}(T)-L^*_n(T))}{4p^3}\Bigg]
 \geqslant 1-\exp(C_{\lambda_0,\sigma_0,p}t).$$
\end{lemma}

All proofs of this subsection are in Appendix B.
Since $\maS_T$ includes the optimal predictor for the $\lin$ model, Lemma~\ref{lem:lin_step_est:cond} follows  immediately.

\begin{lemma}\label{lem:lin_step_est:cond}
For any $\Xi_{k-1}$, let $L_n^{k-1}(T)|_{\Xi_{k-1}}$ be the squared error in node $T$ with $t$ cases given $\Xi_{k-1}$. Then there exists $C_{\lambda_0,\sigma_0,p}>0$ such that
\begin{align*}
 & P_{\boldsymbol X,\varepsilon}\Bigg[\max_{\maS_T\sim\Xi_{k}|\Xi_{k-1}}\left(\Delta_{\mbox{\tiny LIN}}\right)\geqslant \frac{\lambda_0(L_n^{k-1}(T)|_{\Xi_{k-1}}-L^*_n(T))}{4p},\;\;\; \forall \Xi_{k-1}\Bigg]\\
 \geqslant& P_{\boldsymbol X,\varepsilon}\left[\lambda_{min}(\tilde {\boldsymbol X}_T^\top\tilde {\boldsymbol X}_T)>\lambda_0, \;\;\; \forall T\in \maT_{k-1}, \forall \Xi_{k-1}\right]\geqslant1-\exp(-C_{\lambda_0,\sigma_0,p}t)\,.
\end{align*}
\end{lemma}

Next, this yields a bound for an entire tree.

\begin{theorem} \label{theorem:recursion2}
Assume the data is generated by the linear model $Y\sim\boldsymbol X\beta + \varepsilon$ with $n$ cases. Under Conditions 1 and 2, the difference between the training loss $L^k_n$ of a \textit{random PILOT tree} (depending on $\Xi_K$) at depth $K$ and the training loss $L^*_n$ of least squares regression satisfies
\begin{equation}
\maE_{\boldsymbol X,\varepsilon,\Xi_k}[L^K_n-L^*_n] \leqslant \gamma^K\sqrt{\maE_{\boldsymbol X,\varepsilon,\Xi_k}[(L^0_n-L^*_n)^2]}+\mathcal{O}(\log(n)/n^{\delta})
\end{equation}
where 
$$\gamma:=1-\frac{q\lambda_0}{4p^2}\;.$$
\end{theorem}

Now we can prove the fast convergence rate of RaFFLE on linear data.

\begin{theorem}[Fast convergence on linear data] \label{theorem:rate}
Assume the conditions of Theorem~\ref{theorem:recursion2} hold and that $|Y|$ is a.s. bounded. Let $K_n= log_\gamma(n)$. Then we have for any $0<\delta<1$ that
\begin{equation}
  \maE_{\maD_n,\varepsilon,\Xi_{K_n}}[||\hat f(\mathcal {R}_{M,K_n}) - \boldsymbol X\beta||^2] \leqslant \mathcal{O}\left(\frac{\log(n)}{n^{\delta}}\right)\;.
\end{equation}
\end{theorem}

\begin{figure}[ht]
    \centering
    \includegraphics[width=0.95\textwidth]{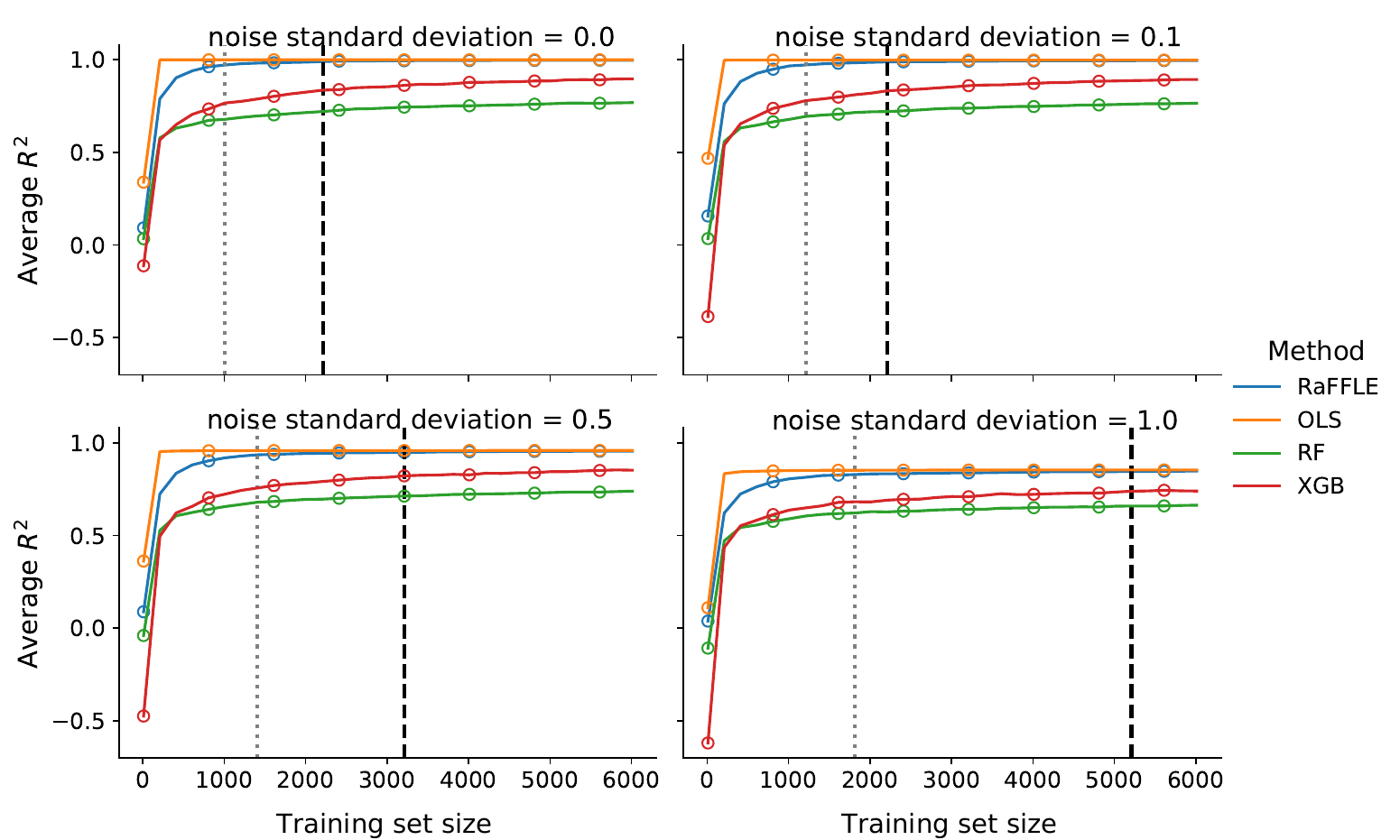}
    \caption{Average $R^2$-scores over 5 runs on simulated linear data, for increasing training set size.}
    \label{fig:linear_convergence}
\end{figure}

To illustrate RaFFLE's favorable convergence rate on linear data, we ran the following simulation. We generate a random feature matrix $X$ with $n = 8000$ cases, $p = 20$ features and an effective rank of 16. The target variable $y$ is subsequently calculated using random coefficients drawn from a uniform distribution in the range $[0, 100]$, plus Gaussian noise with mean zero and a standard deviation of 0, 0.1, 0.5 and 1. From these data we take out a fixed test set of size 2000 and keep it separate. Finally, we let the number of training cases vary between 10 and 6000 in steps of size 200 and fit Ordinary Least Squares regression (OLS), RaFFLE, the classical random forest (RF), and XGBoost (XGB). We validate the performance on the test set using the $R^2$-score. This is repeated 5 times using different random seeds. The $R^2$-scores are then averaged over these 5 repetitions.
Figure \ref{fig:linear_convergence} summarizes the results of this experiment.

The first vertical dashed line indicates the point where RaFFLE reaches 97\% of the OLS $R^2$-score, and the second line where RaFFLE reaches 99\%. We see that RaFFLE needs substantially less data to approximate the OLS solution to a given degree than the classical RF and XGBoost. The latter do not even reach the 97\% for a training set size of 6000.

The analysis in this section demonstrates that RaFFLE not only inherits the consistency of its PILOT base learners but also its fast convergence rate on linear data. These results highlight the  reliable performance of RaFFLE across a wide range of regression tasks. The flexibility of RaFFLE makes it a strong candidate for modeling both linear and nonlinear data structures.

\subsection{Time and space Complexity}
Since PILOT has the same time and space complexity as CART as shown in Section 3.3 of \citep{raymaekers2024pilot}, RaFFLE also has the same complexity as the original CART forest. In particular, the presorting procedure on the full dataset requires $\mathcal{O}(n p\log(n))$ time. For each tree, it requires $\mathcal{O}(n)$ time to get one bootstrap sample. At each node, the time complexity of selecting a model is $\mathcal{O}(n q)$ which dominates the $\mathcal{O}(q)$ time for selecting a random set of $q\leqslant p$ predictor variables. It also dominates the  time for splitting nodes after model selection, which is at most $\mathcal{O}(n)$. The number of nodes in a tree is at most $n^K$. Finally, the procedure is repeated $M$ times to build the forest, which in total gives $\mathcal{O}(n p\log(n) + M2^Knq)$. 

The space complexity follows directly, and equals $\mathcal{O}(np + M2^Kp)$. The first term is the size of the dataset, and the second is for storing the fitted parameters of the trees, which includes split points and linear coefficients. 

\section{Empirical study}\label{pilot_sec:results}
\subsection{Datasets and methods}
Now that the theoretical guarantees on RaFFLE are established, we investigate its properties empirically. We compare RaFFLE with both classical and state-of-the-art regression methods, on datasets of different types.

We benchmark RaFFLE on a total of 136 regression datasets from 2 sources: the UCI Machine Learning Repository (UCI) \cite{uciml} and the Penn Machine Learning Benchmark (PMLB) \cite{Olson2017PMLB, romano2021pmlb}. From both sources we downloaded all datasets that are available through their respective Python APIs (\url{https://github.com/uci-ml-repo/ucimlrepo} and \url{https://github.com/EpistasisLab/pmlb}) and are labeled as \textit{Regression} tasks. We found that some of the datasets labeled as \textit{Regression} actually had a non-numeric target variable, so we removed them from the selection. We also removed the three largest regression datasets from PMLB (1191, 1195 and 1595), each with 1 million observations, as running the grid search on them was too computationally expensive. We also removed the Forest Fire dataset \cite{forest_fires_162} (UCI 162) because all methods in our study had a negative $R^2$ on this dataset, due to substantial outliers \cite{Cortez2007ADM}. 

We did very little preprocessing on the datasets: (1) features consisting of dates were dropped, (2) columns with more than 50\% of missing values were dropped, 
and (3) in the Online News Popularity dataset \cite{online_news_popularity_332} we log-transformed the skewed target variable.\\

We compare RaFFLE with the following competitors:
\begin{itemize}
\itemsep -1pt
    \item \textbf{CART}: a single CART decision tree \cite{cart};
    \item \textbf{PILOT}: a single PILOT decision tree \cite{raymaekers2024pilot};
    \item \textbf{RF}:  a classical random forest of CART decision trees \cite{breiman2001random};
    \item \textbf{XGB}: the popular XGBoost method \cite{xgboost}; 
    \item \textbf{Lasso}: $\ell_1$-regularized linear regression;
    \item \textbf{Ridge}: $\ell_2$-regularized linear regression.
\end{itemize}
For CART, RF, Lasso and Ridge we use the scikit-learn \cite{scikit-learn} implementations, and for XGB we use the official python API \url{https://xgboost.readthedocs.io/en/stable/python/index.html}\,.
For CART and PILOT we only consider the default parameters. For Lasso and Ridge the parameter controlling the level of regularization is tuned by searching over a grid of 100 values. Finally, for RF, RaFFLE 
and XGB we perform a grid search on the hyperparameters listed in Table \ref{tab:grid_search_parameters}, where \textsc{n\_features\_node} is the fraction of the total number of features $p$ that is drawn at each level of each tree. For \textsc{n\_features\_node} $=0.7$ this is 70\%, and when it is set to 1 all features are used each time, as in the default random forest for regression in scikit-learn.  

As both RF and XGBoost are often used out-of-the-box without much tuning of hyperparameters, we include a default version of RaFFLE alongside the tuned RaFFLE. In this way we can investigate the sensitivity of RaFFLE to the hyperparameters. Moreover, it allows for a fair comparison between RaFFLE and RF, given that the cross-validated RaFFLE has an extra tuned parameter $\alpha$. The default RaFFLE, which we will denote by \textbf{dRaFFLE}, has parameters \textsc{alpha} = 0.5, \textsc{n\_features\_node} = 1.0 and \textsc{max\_depth} = 20. All other RaFFLE parameters (for both the tuned and the default version, see Algorithm \ref{algo:RaFFLE} for an overview) are set to the following default values: \textsc{n\_estimators} = 100, \textsc{n\_features\_tree} = 1, \textsc{max\_model\_depth} = 100, \textsc{min\_sample\_fit} = 10, \textsc{min\_sample\_alpha} = 5, and \textsc{min\_sample\_leaf} = 5. 

\begin{table}[!ht]
\centering
\caption{Overview of grid search parameters}
\begin{tabular}{|>{\raggedright\arraybackslash}p{3.8cm}|>{\raggedright\arraybackslash}p{3.8cm}|>{\raggedright\arraybackslash}p{3.8cm}|}
\hline
\textbf{Parameter name} & \textbf{Parameter values} & \textbf{Used by} \\
\hline
\textsc{alpha} & \{0.01, 0.5, 1\} & RaFFLE \\
\hline
\textsc{n\_features\_node}$^1$  & \{0.7, 1\} & RF, RaFFLE, XGB \\
\hline
\textsc{max\_depth} & \{6, 20, None$^2$\} & RF, RaFFLE, XGB \\
\hline
\end{tabular}
\label{tab:grid_search_parameters}  

\vspace{1mm}
\footnotesize{1. for RF the parameter is called \textsc{max\_features}, for XGB \textsc{colsample\_bynode}}.\\
\footnotesize{2. \textsc{max\_depth} = None is only used in RF as it is the scikit-learn default.}
\end{table}

\subsection{Results}

In order to compare the methods, we compute the average five-fold cross-validation $R^2$ for each of the methods. In other words, for each fold we train on 80\% of the data, and then compute the $R^2$ on the holdout part containing the remaining 20\% of the data. Then we compute the average of these $R^2$ values over the five folds. This yields a performance measure for all the different methods. 

While the resulting performance measures are comparable for a given dataset, we need to adjust the performance measure for the different degrees of difficulty across datasets. Ranks are often used for this, but they do not capture the amount by which the performance differs between methods. Instead, we carry out the following normalization. For each dataset, we divide the average $R^2$ values by the highest of these values attained by the various methods on the same dataset. The resulting performance measure lies between 0 and 1 (we clip negative $R^2$ values to 0), where 1 indicates it is the best performer, and a value smaller than 1 quantifies the performance of the method relative to the best method on that dataset. The resulting relative $R^2$ values allow comparisons across datasets.

Tables \ref{tab:empirical_results_p1}, \ref{tab:empirical_results_p2} and \ref{tab:empirical_results_p3} in Section \ref{app:fullsimulationresults} of the Supplementary Material show the detailed results of the benchmarks on all datasets. The summary statistics are shown in Table \ref{tab:empirical_results_summary}, from which we can draw several conclusions. RaFFLE with tuned hyperparameters achieves an average relative $R^2$-score of 0.99 with a standard deviation of 0.02, and substantially outperforms all other methods. Interestingly, dRaFFLE performed second best at 0.96, in spite of its fixed hyperparameters. This indicates that the default parameters in dRaFFLE already give a reasonably good performance, making it suitable as an out-of-the-box method that is faster than the full RaFFLE. Next in line are PILOT, RF and XGB with performances around 0.90\,. Finally, we have CART, Lasso and Ridge with average performance about 40\% below the best method per dataset, and whose performance is much more variable.

\begin{table}[ht]
\caption{Summary statistics of the relative $R^2$ values.}
\label{tab:empirical_results_summary}
\centering
\begin{tabularx}{\textwidth}{lrrrrrrrr}
\toprule
    &          CART &         PILOT &            RF &   
    RaFFLE &  dRaFFLE&           XGB &         Ridge &  
    Lasso \\
\midrule
  mean &          0.64 &          0.91 &          0.92 & 
  \textbf{0.99} & 0.96 &           0.90 &         0.59 & 
  0.58 \\
  std &          0.28 &          0.15 &          0.09 & 
  \textbf{0.02} &      0.10 &      0.15 &        0.33 &
  0.31 \\
\bottomrule
\end{tabularx}
\end{table}

\begin{figure}[!ht]
\centering\includegraphics[width=0.8\textwidth]{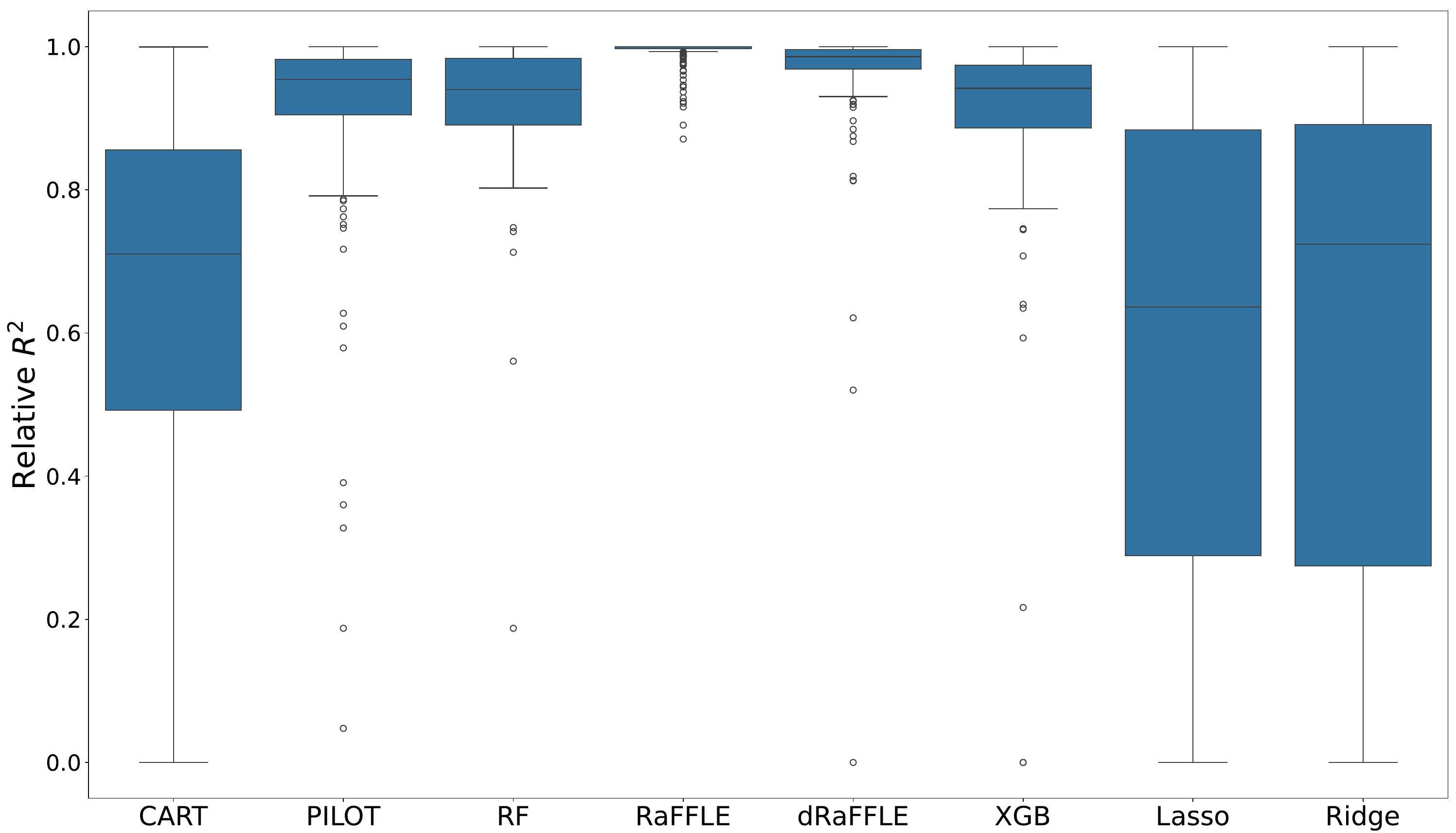}
\caption{Boxplots of relative $R^2$-scores by method}
\label{fig:boxplot_overall}
\end{figure}

Figure \ref{fig:boxplot_overall} shows boxplots of the relative $R^2$ values over all datasets, which confirm the initial conclusions. RaFFLE stands out with the highest relative $R^2$ values, which were always above 80\% and very often above 95\%. dRaFFLE is the second best performer, followed by PILOT, RF and XGB. Note that this instance of PILOT is the default version without tuning any parameters, whereas they were tuned for RF and XGBoost. Finally, we have CART, Lasso and Ridge that do well on some datasets, but have a median performance of about 70\% of the best method, and a substantial number of datasets where the performance is quite poor relative to the other five methods.

\begin{figure}[!ht]
\centering
\includegraphics[width=0.7\textwidth]{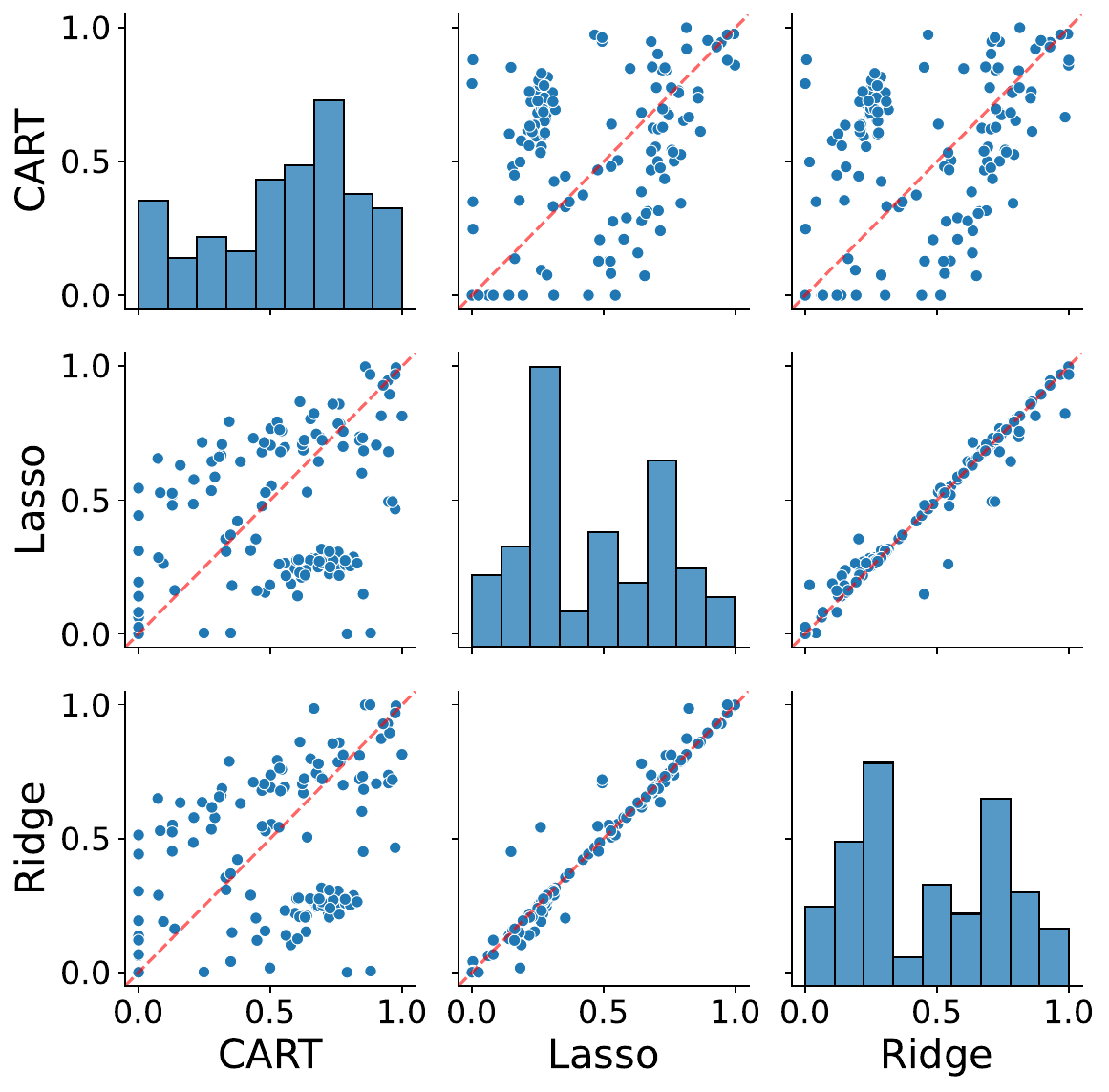}
\caption{Pairs plot of raw $R^2$ values of CART and the linear methods Lasso and Ridge}
\label{fig:pairplot_zoom}
\end{figure}

Another analysis of the empirical results takes the presence of strong linear patterns in the data into account. Sometimes CART performs well, and sometimes the linear methods Lasso and Ridge do well. Given the fundamental difference in the structures that these methods aim to fit, we can expect that their relative performance varies a lot over the different datasets. This is indeed confirmed when plotting the unstandardized $R^2$ values of CART against those of Lasso and Ridge, as shown in Figure \ref{fig:pairplot_zoom}. We see that the performances of Lasso and Ridge are relatively similar, and both differ a lot from the performance of CART.

\begin{figure}[!ht]
\centering
\includegraphics[width=1.0\textwidth]{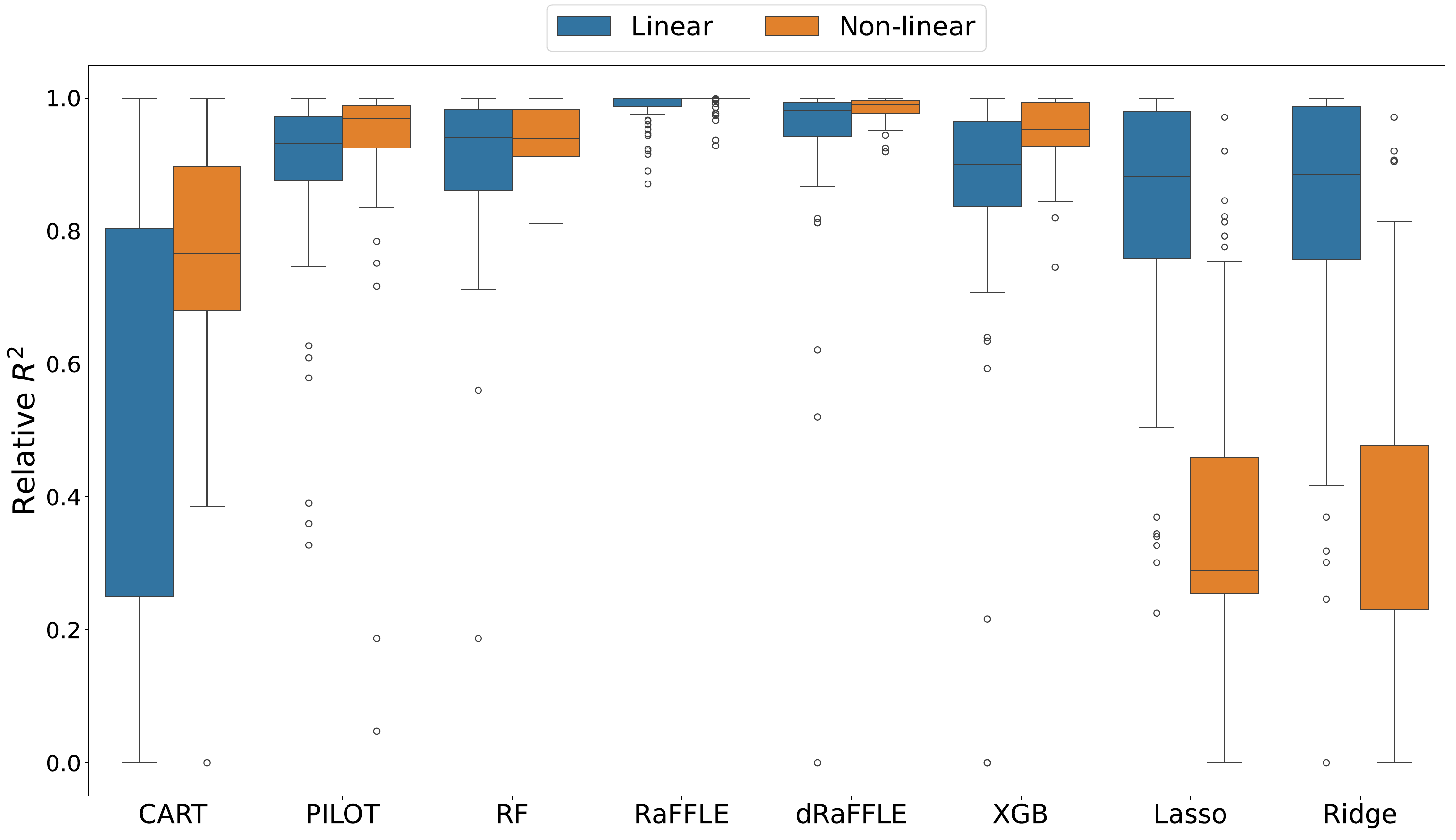}
\caption{Boxplots of relative $R^2$-scores by method and `type' of dataset. The so-called `linear datasets' (blue boxplots) are those where Lasso or Ridge performed better than CART. On the remaining `nonlinear datasets' (orange boxplots) CART performed better than Lasso and Ridge.}
\label{fig:boxplots_lin_vs_nonlin}
\end{figure}

We can thus try to categorize the datasets as `linear' and `nonlinear'. The so-called `linear datasets' are those where Lasso or Ridge outperform CART, whereas the others are considered `nonlinear datasets' on which CART outperforms. We can then compare the performance of the methods on both types separately. Figure \ref{fig:boxplots_lin_vs_nonlin} shows the same relative $R^2$ values as Figure \ref{fig:boxplot_overall}, but now split by dataset type. RaFFLE outperforms on both data types, followed by dRaFFLE. We also see that the linear methods typically performed poorly on the nonlinear datasets, whereas CART often still performed well on linear datasets. The relative performances of the remaining nonlinear methods PILOT, RF, and XGB were fairly similar to each other.

\vspace{-2mm}
\section{Conclusions}\label{pilot_sec:conclusion}

In this paper we introduced RaFFLE, a random forest of linear model trees trained by PILOT. By integrating this  base learner, RaFFLE combines the versatility of random forests with the expressiveness of linear model trees. This hybrid approach enables RaFFLE to handle a wide range of regression problems, from datasets with complex nonlinear relationships to those with strong linear or piecewise linear structures.

To adapt PILOT for use in random forests, we made targeted modifications to increase the variability between the trees by reducing the regularization in the standalone PILOT's design. To this end we introduced the parameter $\alpha$ to tune the degrees of freedom assigned to different node types, added node-level feature sampling, and disabled the \textsc{blin} node type to speed up the computation. These modifications allow individual trees in the ensemble to overfit slightly in order to increase diversity within the forest, a critical factor in improving overall predictive performance. The resulting RaFFLE algorithm strikes a balance between accuracy and computational efficiency.

Our theoretical analysis shows that RaFFLE is consistent at general additive data. We also show that it attains a better convergence rate on data generated by a linear model. Empirical evaluation on 136 regression datasets from the UCI and PMLB repositories demonstrates RaFFLE’s superior performance compared to CART, the traditional random forest, XGBoost, Lasso and Ridge. RaFFLE not only achieves high accuracy across a broad spectrum of datasets, but also consistently outperforms other methods on both linear and nonlinear tasks. Even a non-tuned default version of RaFFLE performs competitively, underscoring the stability of the approach.

Despite its strengths, RaFFLE has some limitations that merit further investigation. The computational complexity of PILOT nodes, while mitigated by our modifications, can still pose challenges for extremely large datasets. Future work could explore more efficient tree-building algorithms or parallelized implementations to improve scalability. Examining its performance in high-dimensional settings with noisy or sparse features, as well as in domains with a large percentage of missing data, could uncover new opportunities for improvement. Moreover, while RaFFLE performs well in regression tasks, extending the framework to handle classification problems would broaden its applicability.

Another promising avenue is the integration of advanced optimization techniques, such as gradient boosting or adaptive sampling, which could further enhance RaFFLE’s accuracy and speed. Lastly, refining interpretability tools, such as feature importance measures or visualizations tailored to PILOT’s piecewise linear structure, could improve RaFFLE's appeal in domains where transparency is critical.

RaFFLE represents a step forward in the evolution of tree-based ensembles, bridging the gap between traditional decision trees and linear model trees. By leveraging PILOT’s speed and regularization within a random forest framework, RaFFLE offers a unique blend of flexibility and accuracy. This hybrid approach enables it to address a wide range of regression challenges.

\vspace{5mm}
\noindent {\Large \bf Appendix} 

\section*{A. Proofs of the theoretical results in Section~\ref{subsec:consistency}}
\label{app:consistency}

\noindent {\bf Proof of Lemma~\ref{lem:pcon_step_est}}
Let $\widehat\Delta^{k}_{\mbox{\tiny PCON}}(\hat s_p, \hat j_p, T)$ be the largest impurity gain induced by the optimal predictor $\hat j_p\in\{1,\dots,p\}$ and split point $\hat s_p$. Following the reasoning for equation (E.2) in \cite{klusowski2024large}, we know that $\maE_{\Xi_k|\Xi_{k-1}}\left[\max_{j\in\maS_T}\widehat\Delta^{k-1}_{\mbox{\tiny PCON}}(\hat s_j,j,T)\right]\geqslant q\widehat\Delta^{k-1}_{\mbox{\tiny PCON}}(\hat s_p, \hat j_p, T)/p$. The intuition is that the expected gain should be larger than the optimal gain times the probability that the optimal predictor is included in the set of predictors selected by $\Xi_k$, which equals $q/p$. Combining this with the lower bound on $\widehat\Delta_{\mbox{\tiny PCON}}$ in Lemma 3 of \citep{raymaekers2024pilot} proves the result. Note that the $2B$ in the denominator comes from the fact that a random PILOT tree truncates at $[-B,B]$, and therefore $\sum_{j=1}^{p_n}||\hat f(\maT_k)_j||_{TV}\,=\max \hat f(\maT_k)-\min\hat f(\maT_k)\leqslant 2B$, from which we obtain $\left(\frac{1}{2}||f-\hat f(\maT_k)||_{TV(T)}\right)^2 \leqslant \frac{1}{4}(||f||_{TV}+2B)^2$\,.

\newpage
\noindent {\bf Proof of Lemma~\ref{lem:bic}}

\begin{proof}
The first result follows directly from taking the conditional expectation. For the second result, we note that for each realization of $\Xi_{k}|\Xi_{k-1}$, $\widehat\Delta$ and the selected model are deterministic. Therefore, by Proposition 4 of \citep{raymaekers2024pilot}, it holds that for $\alpha > 0$ we have $\widehat\Delta\geqslant\frac{(\nu_{\lin}-1)}{(\nu_{\plin}-1)}\widehat\Delta_{\mbox{\tiny PCON}} = \frac{1}{4}\widehat\Delta_{\mbox{\tiny PCON}}$. When $\alpha = 0$ the BIC is equivalent to the MSE, so we have $\widehat\Delta\geqslant\widehat\Delta_{\mbox{\tiny PCON}}$. Therefore, by taking the conditional expectation we prove the result on $E_2$. 
    
\end{proof}

\noindent {\bf Proof of Theorem~\ref{theorem:errorRandomPILOT}}
\begin{proof}
As we saw in Remark \ref{rem:connode}, selecting $\con$ in a node does not necessarily mean that $\con$ will be selected in subsequent steps. But there may be nodes where $\con$ is the overall the best, that is, no other model on any of the predictors achieves a lower BIC. 
We denote these \textit{final} $\con$ nodes by $\con^*$ and separately control the error on them. 
We follow the notations in the proof of Theorem 3 of \citep{raymaekers2024pilot}, but we define the set $C^+_k$ only for those final $\con^*$ nodes. 
In particular, for each realization of $\Xi_K$, $1\leqslant k\leqslant K$, we let $C_k^+:=\{T|T=\textsc{con}^*, T\in \maT_{k-1}, R_{k-1}(T)>0\}$ be the set of nodes on which $\textsc{con}^*$ is fitted before the \mbox{$k$-th}\linebreak step, $R_{C_k^+}:=\sum_{T\in C^+_k}w(T)R(T)$, and $\widetilde{R}_k:=R_k-R_{C_k^+}$. We want to first evaluate the expected value of $\widetilde R_k - \widetilde R_{k-1}$ which is the impurity gain on the non-$\con^*$ nodes at step $k$. The errors $\maE_{\Xi_k}(R_{C_K^+})$ in the $\con^*$ nodes will be added later. Here we consider two specific situations for each $T$: 

\noindent 1. The optimal predictor $\hat j_p$ among all predictors is included in $\maS_T$. The probability is $q/p$ in each node, as in the proof of Lemma \ref{lem:pcon_step_est}.

\noindent 2. $\hat j_p$ is not included in $\maS_T$ (this probability is $1-q/p$ in each node). In this case, either other suboptimal non-$\con$ models are selected or a $\con$ model is selected which might not necessarily be $\con^*$. 

In the first situation, we have by the proof in \citep{raymaekers2024pilot} that
\begin{equation*}
        \widetilde R_k\leqslant \widetilde R_{k-1} - \frac{1}{F}\widetilde R^2_{k-1}
\end{equation*}
where the constant $F = 4(||f||_{TV}+2B)^2$ comes from the second result in Lemma \ref{lem:bic} which controls the lower bound of the ratio between the gain of $\pcon$ (on all predictors) and the selected model (on $\hat j_p$). In the second situation, the impurity gain is trivially bounded by zero where the worst case corresponds to a $\con$ fit. We now take the expectation over $\Xi_{k}|\Xi_{k-1}$ to get 
\begin{equation*}
    \widetilde R_{k-1} - \maE_{\Xi_{k}|\Xi_{k-1}}[\widetilde R_k]\geqslant \maE_{\Xi_{k}|\Xi_{k-1}} \! \left[ \sum_{T\in \maT_{k-1}\backslash C^+_{k-1},  \tilde R_{k-1}(T)\geqslant0}\mathbbm 1_{\{\hat j_p\in \maS_T\}}\frac{w(T)}{F}\widetilde R^2_{k-1}(T)\right] \geqslant \frac{q}{pF}\widetilde R^2_{k-1}.
\end{equation*}
By taking the expectation with respect to $\Xi_{k-1}$ and applying Jensen's inequality to\linebreak $\maE_{\Xi_{k-1}}[\widetilde R^2_{k-1}]$ we get
 \begin{equation*}
   \maE_{\Xi_k}[\widetilde R_k]\leqslant\maE_{\Xi_k}[\widetilde R_{k-1}] - \frac{q}{pF}(\maE_{\Xi_k}[\widetilde R_{k-1}])^2\,.
\end{equation*}
By an induction argument similar to Lemma 4.1 of \cite{klusowski2024large} we obtain 
\begin{equation*}
    \maE_{\Xi_k}[||Y-\hat f(\mathcal{T}_K)||^2_n]\, \leqslant ||Y-f||^2_n \, + \frac{4p(||f||_{TV}+2B)^2}{q(K+3)} + \maE_{\Xi_k}[R_{C_K^+}]\;.
\end{equation*}
Next we deal with the $\con^*$ nodes. For $\maE_{\Xi_k}(R_{C_K^+})$, we note that for $\forall T\in C_K^+$, we have
    \begin{equation*} \label{eq:conbound}
    \frac{(\maE_{\Xi_k} [R(T)])^2}{FR_0}\leqslant\frac{\maE_{\Xi_k} [R(T)^2]}{FR_0}\leqslant \frac{p\maE_{\Xi_{k}}[\maE_{\Xi_{k+1}|\Xi_k}[\Delta_{\mbox{\tiny PCON}}]]}{qR_0}\leqslant\frac{p}{qt}\Big(1-\exp\Big((\nu_{\mbox{\tiny CON}}-\nu_{\mbox{\tiny PCON}})\frac{\log t}{t}\Big)\Big)
\end{equation*}
where $R_0$ is the initial error before training, and $R(T)$ is the remaining error in $T$. The first inequality follows from Jensen's inequality, the second inequality follows from Lemma \ref{lem:pcon_step_est} which provides an upper bound of the error $R(T)$ using the expected gain of a potential $\pcon$ fit, and the last inequality follows from the first result in Lemma \ref{lem:bic}. Note that by our definition, $\con^*$ is the best model considering all predictors, therefore the conditional expectation in that first result becomes an unconditional expectation, which justifies the last inequality. Note that we have controlled the expected error in a $\con^*$ node by its number of cases $t$. We then apply the Cauchy-Schwarz inequality to get a bound only regarding $n$,
\begin{equation*}
    \maE_{\Xi_k}(R_{C_K^+})\le\sqrt{2^Kn\log n}\precsim \mathcal{O}\left(\sqrt{\log (n)/n^{(r-1)/r}}\,\right)
\end{equation*}
    by choosing $K=\log_2(n)/r$ with $r>1$.
\end{proof}

\noindent {\bf Proof of Theorem~\ref{theorem:consistency}}

\begin{proof}
As in the proof of Theorem 4.3 in  \cite{klusowski2024large}, we will leverage Theorem 11.4, Theorem 9.4, and Lemma 13.1 from \cite{distribution} to show this. For each $m$ we have
\begin{align*}
    P\Big(&\maE_{\Xi_{K_n}}||f-\hat f(\maT_{K_n}^{(m)})||^2\geqslant2(\maE_{\Xi_{K_n}}||Y-\hat f(\maT_{K_n}^{(m)})||^2_n-||Y-f||^2_n)+a+b \Big) \leqslant \\
    &P\Big(\exists f\in\mathcal F,\text{ } ||f-\hat f(\maT_{K_n}^{(m)})||^2\geqslant2(||Y-\hat f(\maT_{K_n}^{(m)})||^2_n-||Y-f||^2_n)+a+b\Big).
\end{align*}
Here $a,b$ are positive numbers that tend to zero as $n\rightarrow\infty$. The above holds because the inequality for the expectation implies the existence of a piecewise linear function and a realization of $\Xi_{K_n}$ for which the inequality holds (i.e., without expectations). Following \cite{klusowski2024large}, we assume that the procedure samples a subset 
$\maD_{\maI_m}\subset \maD_n =\{X_1,\dots,X_n\}$ 
with indices $\maI_m\subset \{1,\dots,n\}$
such that $\#\maI_m = n_m$ and 
$\maE_{\maI_m}(1/n_m)\asymp 1/n$ (e.g. with fixed 
size $n/2$). 
Therefore, we can follow the proof of Theorem 1 of \citep{raymaekers2024pilot} to deduce that on $\maD_{\maI_m}$ we have
\begin{equation} \label{oracle}
\begin{split}
\maE_{\Xi_{K_n}}[\,||f-\hat f(\maT_{K_n}^{(m)})||^2]\leqslant &\,\frac{2p_nF}{q_n({K_n}+3)}+\frac{C_3\sqrt{2^{K_n}n_m\log n_m}}{n_m}\\&\,+\frac{C_4\log(n_m p_n\log n_m)2^{K_n+\log(p_n+1)}}{n_m}\,.
\end{split}
\end{equation}
Using our assumption that $\maE_{\maI_m}(1/n_m)\asymp 1/n$ we have 
\begin{equation*}
    \maE_{\Xi,\maD_n}[||f-\hat f(\mathcal {T}^{(m)}_{K_n})||^2]\leqslant \frac{2p_nF}{q_n(K_n+3)}+\frac{C_3'\sqrt{2^{K_n}n\log n}}{n}+\frac{C_4'\log(n p_n\log n)2^{K_n+\log(p_n+1)}}{n}
\end{equation*}
by taking the expectations over all possible $\maD_{\maI_m}$ and $\Xi_K$. Finally, by Jensen's inequality, we can upper bound the error of the forest by the average error of the random PILOT trees to get
\begin{equation*}
    \maE_{\Xi,\maD_n}[||f-\hat f(\mathcal {R}_M)||^2]\leqslant \frac{2p_nF}{q_n(K_n+3)}+\frac{C_3'\sqrt{2^{K_n}n\log n}}{n}+\frac{C_4'\log(n p_n\log n)2^{K_n+\log(p_n+1)}}{n}
\end{equation*}
which tends to zero if $K_n, p_n$, and $q_n$ satisfy the conditions.
\end{proof}

\section*{B. Proofs of the theoretical results in Section~\ref{subsec:convergencelinear}}
\label{app:linear}

\noindent {\bf Proof of Lemma~\ref{lem:lin_step_est}}

\begin{proof}
We re-parameterize $L_n^k(T)=||Y_T-\boldsymbol{\tilde X}_T\tilde\beta_T||^2_t$ and write its gradient as $\nabla L_n^k(T)|_{\tilde\beta = \tilde\beta_T}\,$. By the definition of the gradient, we have for fixed $\tilde{\boldsymbol{X}}_T$ and noise that
\begin{align}\label{eq:lem1:lin_est}
\begin{split}
  \maE_{\Xi_k|\Xi_{k-1}}\left[\frac{nq||\nabla L_n^{k-1}(T)|_{\tilde\beta = \tilde\beta_T}||_{\infty}}{2p}\right]&=\maE_{\Xi_k|\Xi_{k-1}}\left[q\left|\left|r^\top \tilde{\boldsymbol X}_T\right|\right|_\infty/p\right]\\
  &=\maE_{\Xi_k|\Xi_{k-1}}\left[q\max_{j\in\{1\dots p\}}\{|r^\top \tilde X^{(j)}_T|\}/p\right]\\
  &\leqslant\maE_{\Xi_k|\Xi_{k-1}}\left[\max_{j\in\maS_T}\frac{|r^\top X^{(j)}_T|}{\sqrt n\hat\sigma_j^u}\right]\\
  &\leqslant \maE_{\Xi_k|\Xi_{k-1}}\left[\max_{j\in\maS_T}\frac{|r^\top (X^{(j)}_T-\overline X^{(j)}_T)|}{\sqrt n\hat\sigma_j}\right]\\
  &=\maE_{\Xi_k|\Xi_{k-1}}\left[\sqrt{n\Delta_{\mbox{\tiny LIN}}}\right]
\end{split}
\end{align}
where $r$ denotes the residuals $Y_T-\tilde \bX_T\tilde\beta_T$. Note that $\nabla L_n^{k-1}$ is deterministic given $\tilde{\boldsymbol{X}}_T$, $\Xi_{k-1}$ and the noise. The first inequality follows from the fact that the optimal predictor is included in the sampled predictors with probability $q/p$ (see also Lemma \ref{lem:pcon_step_est}), the second inequality follows from the fact that the residuals of the response on $T$ have zero mean, and the last equality follows from Lemma 1 of \citep{raymaekers2024pilot} which gives an explicit expression of $\Delta_{\lin}$. Therefore, we have that
\begin{align*}
 &P_{\boldsymbol X_T}\Bigg[\maE_{\Xi_k|\Xi_{k-1}}\left(\Delta_{\mbox{\tiny LIN}}\right)\geqslant 
 \frac{q^2\lambda_0(L_n^{k-1}(T)-L^*_n(T))}{4p^3}\Bigg]\\\geqslant &P_{\boldsymbol X_T}\left[\frac{n||\nabla L_n^{k-1}|_{\tilde\beta = \tilde\beta_T}||^2_{\infty}}{4}\geqslant \frac{\lambda_0(L_n^{k-1}(T)-L^*_n(T))}{4p}\right]\\
  \geqslant &P_{\boldsymbol X_T}\left[\lambda_{min}(\tilde {\boldsymbol X}_T^\top\tilde {\boldsymbol X}_T)>\lambda_0\right]
\end{align*}
where the first inequality follows from equations in (\ref{eq:lem1:lin_est}). The second inequality follows from the fact that $L_n^*(T)$ and $ L_n^{k-1}(T)$ are deterministic given $\Xi_{k-1}$, $\tilde{\boldsymbol{X}}_T$ and the noise. Since $\Xi_{k-1}$ only affects the node (hyperrectangle) $T$ in which $\tilde{\boldsymbol{X}}_T$ lies, and Conditions 1 and 2 hold uniformly for any $T$, we can apply the reasoning in the proof of Lemma 4 in \citep{raymaekers2024pilot}.
\end{proof}

\noindent {\bf Proof of Theorem~\ref{theorem:recursion2}}

\begin{proof}
Following the proof of Theorem 4 in \citep{raymaekers2024pilot} and taking the expectation 
with respect to $\Xi_k$ we have with probability at least $\sum_{T\in \maT_k}\exp(-C_{\lambda_0,\sigma_0,p}t)$ under $\boldsymbol{X}, \varepsilon$ that
\begin{align*}
    \maE_{\Xi_{k-1}}[ L_n^{k-1}-L^*_n] - &\maE_{\Xi_k}[(L^{k}_n - L^*_n)]
    \geqslant\maE_{\Xi_{k}}\left[\mathbbm 1_{\{\hat j_p\in\maS_T\}}\sum_{T\in \maT_{k-1}}w(T)\Delta_{\mbox{\tiny LIN}}(T)\right]\\
    &\geqslant \frac{q}{p}\maE_{\Xi_{k-1}}\left[\frac{\lambda_0}{4p}\maE_{\Xi_k|\Xi_{k-1}}\left(\sum_{T\in\maT_{k-1}}w(T)(L^{k-1}_n(T) - L^*_n(T))\right)\right]\\
    & \geqslant  \frac{q\lambda_0}{4p^2}\maE_{\Xi_{k-1}}[L^{k-1}_n-L^*_n].
\end{align*}
Here the first inequality follows from the proof of Theorem \ref{theorem:errorRandomPILOT} where we discuss whether the optimal predictor $\hat j_p$ is included. The second inequality follows from our Lemma~\ref{lem:lin_step_est:cond} and (i)-(iv) of the  proof of Theorem 4 in \citep{raymaekers2024pilot}. Note that $\hat j_p$ need not be the optimal predictor of $\lin$, and the optimal model must have a larger impurity gain than the optimal gain of the $\lin$ model by the BIC criterion. The final inequality follows from (v) of that proof. Therefore,
\begin{equation*}
    \maE_{\Xi_{k}}[L^{k}_n-L^*_n]\leqslant\maE_{\Xi_{k-1}}[L^{k-1}_n-L^*_n]\left(1-\frac{q\lambda_0}{4p^2}\right).
\end{equation*}
Note that here $1-\frac{q\lambda_0}{4p^2}\geqslant1-\frac{\lambda_0}{4p}>0$.

Let $G:=\{(\boldsymbol{X},\varepsilon)|\forall \Xi_K, T\in \maT_k, 1\leqslant k\leqslant K, \lambda_
{min}(\tilde{\boldsymbol X}_T^\top\tilde{\boldsymbol X}_T)\geqslant\lambda_0\}$. By Lemma~\ref{lem:lin_step_est:cond} and a union bound on $K$ (e.g.\ choosing $K\asymp\log n$), we can derive as in \citep{raymaekers2024pilot} that\linebreak $P(G^c)\precsim \exp(-C'_{\lambda_0,\sigma_0,p}n^{\delta})$. We then have for sufficiently large $n$ that
\begin{align*}
\maE_{\boldsymbol X,\varepsilon,\Xi_K}&[L^K_n-L^*_n]=\maE_{\boldsymbol X,\varepsilon}[\mathbbm 1_{G}\maE_{\Xi_{K}}[L^K_n-L^*_n]+\mathbbm 1_{G^c}\maE_{\Xi_{K}}[L^K_n-L^*_n]\\
&\leqslant\sqrt{P(G)}\sqrt{\maE_{\boldsymbol X,\varepsilon}[(\maE_{\Xi_{k}}[L^K_n-L^*_n])^2]} +\sqrt{P(G^c)}\sqrt{\maE_{\boldsymbol X,\varepsilon}[(\maE_{\Xi_{K}}[L^0_n-L^*_n])^2]}\\
&\leqslant \Big( 1-\exp(-C'_{\lambda_0,\sigma_0,p}n^{\delta}) \Big)^{1/2}\left(1-\frac{q\lambda_0}{4p^2}\right)^{K}\sqrt{\maE_{\boldsymbol X,\varepsilon,\Xi_K}[(L^0_n-L^*_n)^2]}\\
&\quad + \exp(-C''_{\lambda_0,\sigma_0,p}n^{\delta})\sqrt{\maE_{\boldsymbol X,\varepsilon,\Xi_K}[(L^0_n-L^*_n)^2]}\\
&\leqslant\left(1-\frac{q\lambda_0}{4p^2}\right)^{K}\sqrt {\maE_{\boldsymbol X,\varepsilon,\Xi_K}[(L^0_n-L^*_n)^2]}+\mathcal{O}(1/n)
\end{align*}
where the second inequality follows from Jensen's inequality.

It remains to control the error in the $\con^*$ nodes. For given $\boldsymbol X,\varepsilon$ we have that if an extra $\lin$ is fitted to the actual $\con$ node, then
\begin{equation*} 
    \frac{\maE_{\Xi_{K+1}|\Xi_{K}}[\Delta_{\mbox{\tiny lin}}]}{R_0}\leqslant\frac{1}{t}\Big(1-\exp\Big((\nu_{\mbox{\tiny CON}}-\nu_{\mbox{\tiny LIN}})\frac{\log t}{t}\Big)\Big)
\end{equation*}
using the first result of Lemma~\ref{lem:bic}. Next, we use Lemma~\ref{lem:lin_step_est} and a similar union bound as we did for $G$ to derive that with probability at least $1-\exp(C_{\lambda_0,\sigma_0,p}t)$ it holds that
\begin{equation*}
   \frac{q^2\lambda_0(L_n^{K}(T)|_{\Xi_{K}}-L^*_n(T))}{4p^3}\leqslant\frac{R_0}{t}\Big(1-\exp\Big((\nu_{\mbox{\tiny CON}}-\nu_{\mbox{\tiny LIN}})\frac{\log t}{t}\Big)\Big),
   \;\;\; \forall \Xi_{K}.
    \end{equation*}
We denote such an event by $H$. Moreover, noting that $\nu_\con\leqslant\nu_\lin$ we have by the integrability of $\varepsilon$ that
\begin{align*}
    \maE_{\boldsymbol X, \varepsilon,\Xi_K}[\mathbbm 1_{H} (L_n^K(T)-L^*_n(T))]&\precsim \maE_{\boldsymbol X,\varepsilon}[Y^2]\maE_{\boldsymbol X, \varepsilon,\Xi_K}{\Big(1-\exp\Big((\nu_{\mbox{\tiny CON}}-\nu_{\mbox{\tiny LIN}})\frac{\log t}{t}\Big)\Big)}\\
    &\precsim\maE_{\boldsymbol X, \varepsilon,\Xi_K}\left({\frac{(\nu_{\mbox{\tiny LIN}}-\nu_{\mbox{\tiny CON}})\log t}{t}}\right).
\end{align*}
Let us denote by $\maT_K^{\textsc{con}^*}$ all the $\textsc{con}^*$ nodes in the tree up to depth $K$. Then, by an argument similar to the case without the $\con^*$ nodes, we control the expectation of the weighted sum of the errors on $\con^*$ nodes by
\begin{align*}
\maE_{\boldsymbol X,\varepsilon,\Xi_K}\Big[\sum_{T\in \maT_K^{\textsc{con}^*}}w(T)&(L_n^K(T)-L^*_n(T))\Big]\\ 
&\leqslant
\maE_{\boldsymbol X,\varepsilon}\Big[\mathbbm 1_{H^c}\sum_{T\in \maT_K^{\textsc{con}^*}}\maE_{\Xi_K}[w(T)(L_n^K(T)-L^*_n(T))]\Big]\\
&\;\;\;\;+\maE_{\boldsymbol X,\varepsilon}\Big[\mathbbm 1_{H}\sum_{T\in \maT_K^{\textsc{con}^*}}\maE_{\Xi_K}[w(T)(L_n^K(T)-L^*_n(T))]\Big]\\
&\leqslant \mathcal{O}\Big(\frac{1}{n}\Big) +\maE_{\boldsymbol X,\varepsilon,\Xi_K}\Big[\mathbbm 1_{H}\sum_{T\in \maT_K^{\textsc{con}^*}}\frac{(\nu_{\mbox{\tiny LIN}}-\nu_{\mbox{\tiny CON}})\log(t)}{n}\Big]\\
&\leqslant \mathcal{O}\Big(\frac{1}{n}\Big) + \mathcal{O}\left(\frac{N_{leaves}\log(n)}{n}\right)\\
&\leqslant \mathcal{O}\Big(\frac{1}{n}\Big) + \mathcal{O}\left(\frac{\log(n)}{n^\delta}\right),
\end{align*}
where the last inequality follows from $N_{leaves}\leqslant n^{1-\delta}$ by Condition 1.

We then combine the two estimations using the same reasoning as in the proof of Theorem 4 in \citep{raymaekers2024pilot}, which concludes the proof.
\end{proof}

\noindent {\bf Proof of Theorem~\ref{theorem:rate}}

\begin{proof}
Using an argument similar to the proof of Theorem 5 in \citep{raymaekers2024pilot} and equation (\ref{oracle}), we first have for any $m$ on $\maD_{\maI_m}$ that
\begin{equation*}
        \maE_{\maD_{I_m},\varepsilon,\Xi_{K_n}}[||\hat f(\mathcal {T}_{K_n}^{(m)}) - \boldsymbol X\beta||^2] \leqslant \mathcal{O}\left(\maE_{I_{m}}\left(\frac{\log(n_m)}{n_m^{\delta}}\right)\right).
\end{equation*}
Since $0 < \delta < 1$ we can apply Jensen's inequality to the concave function $x^\delta$ which yields $\maE_{\maI_m}(1/n_m^\delta)\leqslant(\maE_{\maI_m}(1/n_m))^\delta\precsim1/n^\delta$, and therefore 
\begin{equation*}
        \maE_{\maD_{I_m},\varepsilon,\Xi_{K_n}}[||\hat f(\mathcal {T}_{K_n}^{(m)}) - \boldsymbol X\beta||^2] \leqslant \mathcal O\left(\frac{\log(n)}{n^\delta}\right).
\end{equation*}
Taking the expectation on $\maD_{\maI_m}$ and using Jensen's inequality as in the proof of Theorem \ref{theorem:consistency}  makes the convergence rate hold for the RaFFLE prediction under $\maD_n$.
\end{proof}

\section*{C. Full simulation results per dataset}\label{app:fullsimulationresults}

\renewcommand{\arraystretch}{0.8}
\begin{table}[ht]
\caption{Average $R^2$ score divided by highest average $R^2$ score per dataset (I/III)}
\label{tab:empirical_results_p1}
\centering
\footnotesize
\begin{tabularx}{\textwidth}{llrrrrrrrr}
\toprule
source &  id &          CART &         PILOT &            RF &            RaFFLE &  \makecell{RaFFLE \\Default} &           XGB &         Ridge &         Lasso \\
\midrule
  PMLB & 192 &          0.46 &          0.87 &          0.91 &  \textbf{1.0} &  \textbf{1.0} &          0.64 &          0.89 &          0.89 \\
  PMLB & 195 &          0.87 &          0.91 & \textbf{0.98} &  \textbf{1.0} & \textbf{0.99} & \textbf{0.97} &          0.89 &          0.89 \\
  PMLB & 197 & \textbf{0.98} & \textbf{0.99} &  \textbf{1.0} &  \textbf{1.0} &  \textbf{1.0} &  \textbf{1.0} &          0.73 &           0.5 \\
  PMLB & 201 & \textbf{0.98} &          0.89 &  \textbf{1.0} &          0.93 &          0.92 &  \textbf{1.0} &          0.47 &          0.47 \\
  PMLB & 207 &          0.86 &          0.91 & \textbf{0.98} &  \textbf{1.0} & \textbf{0.98} & \textbf{0.97} &          0.89 &          0.89 \\
  PMLB & 210 &          0.71 &          0.83 &          0.92 & \textbf{0.98} & \textbf{0.98} &           0.9 & \textbf{0.99} &  \textbf{1.0} \\
  PMLB & 215 &          0.95 &  \textbf{1.0} &  \textbf{1.0} &  \textbf{1.0} &  \textbf{1.0} &  \textbf{1.0} &          0.74 &          0.74 \\
  PMLB & 218 &           0.5 &          0.89 & \textbf{0.99} &  \textbf{1.0} &  \textbf{1.0} & \textbf{0.96} &          0.06 &           0.0 \\
  PMLB & 225 &          0.51 & \textbf{0.98} & \textbf{0.98} &  \textbf{1.0} &  \textbf{1.0} &          0.93 &          0.54 &          0.54 \\
  PMLB & 227 & \textbf{0.97} & \textbf{0.99} &  \textbf{1.0} &  \textbf{1.0} &  \textbf{1.0} &  \textbf{1.0} &          0.72 &          0.51 \\
  PMLB & 228 &          0.83 &          0.95 & \textbf{0.96} &  \textbf{1.0} &  \textbf{1.0} & \textbf{0.97} &          0.89 &          0.91 \\
  PMLB & 229 &          0.83 &          0.85 &  \textbf{1.0} &  \textbf{1.0} & \textbf{0.98} & \textbf{0.97} &          0.88 &          0.89 \\
  PMLB & 230 &          0.94 &          0.85 &  \textbf{1.0} &  \textbf{1.0} & \textbf{0.97} &  \textbf{1.0} &           0.9 &          0.82 \\
  PMLB & 294 &          0.86 &          0.88 &  \textbf{1.0} & \textbf{0.97} & \textbf{0.96} &  \textbf{1.0} &          0.78 &          0.78 \\
  PMLB & 344 &  \textbf{1.0} & \textbf{0.99} &  \textbf{1.0} &  \textbf{1.0} &  \textbf{1.0} &  \textbf{1.0} &          0.81 &          0.81 \\
  PMLB & 485 &          0.44 &          0.76 &  \textbf{1.0} & \textbf{0.99} &          0.88 & \textbf{0.99} &          0.87 &          0.88 \\
  PMLB & 503 &          0.68 & \textbf{0.96} & \textbf{0.98} &  \textbf{1.0} & \textbf{0.99} & \textbf{0.97} & \textbf{0.96} & \textbf{0.96} \\
  PMLB & 505 & \textbf{0.98} & \textbf{0.99} & \textbf{0.99} &  \textbf{1.0} &  \textbf{1.0} & \textbf{0.99} &  \textbf{1.0} &  \textbf{1.0} \\
  PMLB & 519 &           0.9 &  \textbf{1.0} &          0.94 &  \textbf{1.0} & \textbf{0.99} &           0.9 &  \textbf{1.0} &  \textbf{1.0} \\
  PMLB & 522 &           0.0 &          0.39 &          0.84 &  \textbf{1.0} & \textbf{0.98} &          0.81 &          0.32 &          0.33 \\
  PMLB & 523 &  \textbf{1.0} &          0.95 & \textbf{0.99} &          0.95 &          0.95 &  \textbf{1.0} & \textbf{0.98} &  \textbf{1.0} \\
  PMLB & 527 &          0.86 &          0.93 &          0.89 &          0.94 &          0.94 &          0.84 &  \textbf{1.0} &  \textbf{1.0} \\
  PMLB & 529 &          0.66 & \textbf{0.98} &          0.95 & \textbf{0.99} & \textbf{0.99} &          0.92 &  \textbf{1.0} &  \textbf{1.0} \\
  PMLB & 537 &          0.75 &          0.93 & \textbf{0.97} &  \textbf{1.0} &  \textbf{1.0} & \textbf{0.98} &           0.6 &          0.62 \\
  PMLB & 542 &          0.23 &          0.91 &          0.85 &          0.89 &          0.62 &          0.82 &  \textbf{1.0} &          0.94 \\
  PMLB & 547 &          0.35 &          0.77 &  \textbf{1.0} & \textbf{0.99} & \textbf{0.99} & \textbf{0.97} &          0.81 &          0.81 \\
  PMLB & 556 &          0.92 &          0.05 &  \textbf{1.0} & \textbf{0.98} & \textbf{0.96} & \textbf{0.98} &           0.0 &           0.0 \\
  PMLB & 557 &          0.95 &          0.19 &  \textbf{1.0} & \textbf{0.98} & \textbf{0.97} &          0.95 &          0.01 &           0.0 \\
  PMLB & 560 & \textbf{0.97} & \textbf{0.99} &  \textbf{1.0} &  \textbf{1.0} &  \textbf{1.0} &  \textbf{1.0} &          0.76 &           0.7 \\
  PMLB & 561 & \textbf{0.96} & \textbf{0.99} & \textbf{0.98} &  \textbf{1.0} &  \textbf{1.0} &  \textbf{1.0} &          0.91 &          0.85 \\
  PMLB & 562 & \textbf{0.97} & \textbf{0.99} &  \textbf{1.0} &  \textbf{1.0} &  \textbf{1.0} &  \textbf{1.0} &          0.72 &          0.51 \\
  PMLB & 564 &          0.88 & \textbf{0.99} & \textbf{0.97} &  \textbf{1.0} &  \textbf{1.0} & \textbf{0.99} &          0.75 &          0.75 \\
  PMLB & 573 & \textbf{0.98} & \textbf{0.99} &  \textbf{1.0} &  \textbf{1.0} &  \textbf{1.0} &  \textbf{1.0} &          0.73 &           0.5 \\
  PMLB & 574 &          0.39 &          0.75 &  \textbf{1.0} & \textbf{0.99} & \textbf{0.98} & \textbf{0.99} &           0.0 &          0.01 \\
  PMLB & 579 &          0.53 & \textbf{0.98} &          0.85 &  \textbf{1.0} & \textbf{0.98} &          0.84 &          0.76 &          0.76 \\
  PMLB & 581 &          0.71 &          0.93 &          0.92 &  \textbf{1.0} & \textbf{0.99} &          0.94 &          0.29 &           0.3 \\
  PMLB & 582 &          0.68 & \textbf{0.97} &          0.87 &  \textbf{1.0} & \textbf{0.98} &           0.9 &          0.22 &          0.25 \\
  PMLB & 583 &          0.71 & \textbf{0.98} &          0.91 &  \textbf{1.0} & \textbf{0.99} &          0.94 &          0.26 &           0.3 \\
  PMLB & 584 &          0.63 &          0.94 &          0.88 &  \textbf{1.0} & \textbf{0.99} &          0.91 &          0.24 &          0.26 \\
  PMLB & 586 &           0.8 & \textbf{0.97} &          0.94 &  \textbf{1.0} &  \textbf{1.0} & \textbf{0.96} &          0.28 &          0.28 \\
  PMLB & 588 &          0.76 &          0.95 &          0.92 &  \textbf{1.0} & \textbf{0.99} &          0.95 &          0.23 &          0.28 \\
  PMLB & 589 &           0.8 & \textbf{0.99} &          0.93 &  \textbf{1.0} & \textbf{0.99} &          0.95 &          0.28 &           0.3 \\
  PMLB & 590 &          0.53 & \textbf{0.96} &          0.83 &  \textbf{1.0} & \textbf{0.99} &          0.89 &          0.73 &          0.75 \\
  PMLB & 591 &          0.12 &          0.91 &          0.89 &  \textbf{1.0} & \textbf{0.98} &          0.88 &          0.25 &          0.34 \\
  PMLB & 592 &          0.81 & \textbf{0.98} &          0.93 &  \textbf{1.0} &  \textbf{1.0} & \textbf{0.96} &          0.25 &          0.26 \\
  PMLB & 593 &          0.78 & \textbf{0.99} &          0.93 &  \textbf{1.0} & \textbf{0.99} &          0.95 &          0.29 &           0.3 \\
  PMLB & 594 &           0.0 &          0.94 &          0.87 &  \textbf{1.0} & \textbf{0.99} &          0.75 &           0.0 &           0.0 \\
\bottomrule
\end{tabularx}
\end{table}

\renewcommand{\arraystretch}{0.9}
\begin{table}[ht]
\caption{Average $R^2$ score divided by highest average $R^2$ score per dataset (II/III)}
\label{tab:empirical_results_p2}
\centering
\footnotesize
\begin{tabularx}{\textwidth}{llrrrrrrrr}
\toprule
source &  id &  CART &         PILOT &   RF &   RaFFLE &  \makecell{RaFFLE \\Default} &  XGB & Ridge & Lasso \\
\midrule
  PMLB & 595 &  0.67 & \textbf{0.98} &          0.89 &  \textbf{1.0} &  \textbf{1.0} &          0.93 &  0.75 &  0.76 \\
  PMLB & 596 &  0.72 & \textbf{0.99} &          0.93 &  \textbf{1.0} & \textbf{0.99} &          0.94 &   0.3 &   0.3 \\
  PMLB & 597 &  0.83 & \textbf{0.99} & \textbf{0.96} &  \textbf{1.0} &  \textbf{1.0} & \textbf{0.97} &  0.27 &  0.27 \\
  PMLB & 598 &  0.59 & \textbf{0.97} &          0.86 &  \textbf{1.0} & \textbf{0.99} &          0.91 &  0.74 &  0.74 \\
  PMLB & 599 &  0.85 & \textbf{0.99} & \textbf{0.96} &  \textbf{1.0} &  \textbf{1.0} & \textbf{0.98} &   0.3 &   0.3 \\
  PMLB & 601 &  0.74 & \textbf{0.98} &          0.92 &  \textbf{1.0} & \textbf{0.99} &          0.92 &  0.33 &  0.34 \\
  PMLB & 602 &   0.7 & \textbf{0.98} &          0.88 &  \textbf{1.0} & \textbf{0.99} &          0.88 &  0.24 &  0.24 \\
  PMLB & 603 &  0.28 & \textbf{0.96} &          0.74 &  \textbf{1.0} & \textbf{0.97} &          0.71 &  0.73 &  0.82 \\
  PMLB & 604 &  0.72 & \textbf{0.98} &          0.93 &  \textbf{1.0} &  \textbf{1.0} &          0.95 &  0.26 &  0.26 \\
  PMLB & 605 &  0.49 &          0.91 &          0.87 &  \textbf{1.0} & \textbf{0.97} &          0.91 &  0.33 &  0.36 \\
  PMLB & 606 &  0.79 & \textbf{0.99} &          0.94 &  \textbf{1.0} & \textbf{0.99} & \textbf{0.96} &  0.32 &  0.32 \\
  PMLB & 607 &  0.76 & \textbf{0.97} &          0.93 &  \textbf{1.0} &  \textbf{1.0} &          0.95 &  0.22 &  0.24 \\
  PMLB & 608 &  0.82 & \textbf{0.98} &          0.95 &  \textbf{1.0} &  \textbf{1.0} & \textbf{0.97} &  0.29 &  0.29 \\
  PMLB & 609 &  0.74 & \textbf{0.97} &          0.92 &  \textbf{1.0} &  \textbf{1.0} &          0.95 &  0.77 &  0.77 \\
  PMLB & 611 &  0.54 &  \textbf{1.0} &          0.82 & \textbf{0.97} &          0.94 &          0.82 &  0.17 &  0.17 \\
  PMLB & 612 &  0.83 & \textbf{0.99} & \textbf{0.96} &  \textbf{1.0} &  \textbf{1.0} & \textbf{0.97} &  0.26 &  0.26 \\
  PMLB & 613 &  0.64 &          0.93 &          0.89 &  \textbf{1.0} & \textbf{0.99} &          0.89 &   0.3 &   0.3 \\
  PMLB & 615 &  0.69 &          0.95 &          0.87 &  \textbf{1.0} & \textbf{0.97} &          0.87 &  0.32 &  0.32 \\
  PMLB & 616 &  0.62 &          0.94 &           0.9 &  \textbf{1.0} & \textbf{0.99} &          0.92 &  0.11 &   0.2 \\
  PMLB & 617 &  0.81 &          0.94 &          0.94 &  \textbf{1.0} &  \textbf{1.0} & \textbf{0.97} &  0.23 &  0.23 \\
  PMLB & 618 &  0.74 & \textbf{0.96} &          0.93 &  \textbf{1.0} &  \textbf{1.0} &          0.95 &  0.26 &  0.28 \\
  PMLB & 620 &  0.72 & \textbf{0.99} &          0.91 &  \textbf{1.0} & \textbf{0.99} &          0.94 &  0.26 &  0.27 \\
  PMLB & 621 &  0.35 &          0.95 &          0.71 &  \textbf{1.0} &          0.87 &          0.74 &  0.77 &   0.8 \\
  PMLB & 622 &  0.75 & \textbf{0.97} &          0.93 &  \textbf{1.0} & \textbf{0.99} &          0.95 &  0.27 &  0.29 \\
  PMLB & 623 &  0.82 & \textbf{0.97} &          0.95 &  \textbf{1.0} &  \textbf{1.0} & \textbf{0.97} &  0.29 &  0.29 \\
  PMLB & 624 &  0.37 &           0.9 &          0.82 &  \textbf{1.0} &           0.9 &          0.85 &  0.79 &  0.79 \\
  PMLB & 626 &   0.7 &          0.92 &          0.91 &  \textbf{1.0} & \textbf{0.97} &          0.95 &  0.17 &  0.26 \\
  PMLB & 627 &  0.75 & \textbf{0.96} &          0.95 &  \textbf{1.0} & \textbf{0.99} &          0.95 &  0.28 &  0.29 \\
  PMLB & 628 &  0.86 & \textbf{0.97} & \textbf{0.96} &  \textbf{1.0} &  \textbf{1.0} & \textbf{0.97} &  0.27 &  0.27 \\
  PMLB & 631 &  0.78 & \textbf{0.98} &          0.94 &  \textbf{1.0} & \textbf{0.99} &          0.94 &  0.29 &  0.29 \\
  PMLB & 633 &  0.51 & \textbf{0.96} &          0.83 &  \textbf{1.0} & \textbf{0.97} &          0.87 &  0.76 &  0.77 \\
  PMLB & 634 &  0.46 &          0.89 &          0.93 &  \textbf{1.0} & \textbf{0.97} &          0.87 &  0.42 &  0.42 \\
  PMLB & 635 &  0.44 & \textbf{0.99} &          0.84 &  \textbf{1.0} & \textbf{0.98} &          0.86 &  0.72 &  0.73 \\
  PMLB & 637 &   0.6 & \textbf{0.97} &          0.86 &  \textbf{1.0} & \textbf{0.97} &          0.91 &  0.15 &  0.23 \\
  PMLB & 641 &  0.77 & \textbf{0.99} &          0.92 &  \textbf{1.0} & \textbf{0.99} &          0.93 &  0.25 &  0.26 \\
  PMLB & 643 &  0.65 & \textbf{0.97} &          0.89 &  \textbf{1.0} & \textbf{0.98} &          0.93 &  0.13 &  0.15 \\
  PMLB & 644 &  0.57 & \textbf{0.96} &          0.83 &  \textbf{1.0} & \textbf{0.96} &          0.86 &  0.02 &  0.21 \\
  PMLB & 645 &  0.66 &          0.92 &           0.9 &  \textbf{1.0} & \textbf{0.98} &          0.92 &  0.22 &  0.25 \\
  PMLB & 646 &  0.77 & \textbf{0.96} &          0.93 &  \textbf{1.0} & \textbf{0.99} &          0.94 &  0.33 &  0.33 \\
  PMLB & 647 &   0.6 & \textbf{0.99} &          0.89 &  \textbf{1.0} & \textbf{0.96} &          0.91 &  0.25 &  0.29 \\
  PMLB & 648 &  0.49 & \textbf{0.97} &          0.84 &  \textbf{1.0} & \textbf{0.97} &          0.88 &  0.22 &  0.39 \\
  PMLB & 649 &  0.67 & \textbf{0.99} &           0.9 &  \textbf{1.0} & \textbf{0.99} &          0.92 &  0.78 &  0.78 \\
  PMLB & 650 &  0.47 & \textbf{0.97} &           0.8 &  \textbf{1.0} & \textbf{0.98} &          0.82 &  0.77 &  0.79 \\
  PMLB & 651 &   0.0 &          0.88 &          0.56 &  \textbf{1.0} &          0.81 &          0.59 &  0.72 &  0.76 \\
  PMLB & 653 &  0.37 &          0.87 &           0.8 &  \textbf{1.0} &          0.95 &          0.81 &   0.8 &  0.82 \\
  PMLB & 654 &  0.59 & \textbf{0.97} &          0.86 &  \textbf{1.0} & \textbf{0.98} &          0.89 &  0.74 &  0.74 \\
  PMLB & 656 &  0.51 &          0.86 & \textbf{0.98} &  \textbf{1.0} &          0.95 &  \textbf{1.0} &  0.21 &  0.26 \\
\bottomrule
\end{tabularx}
\end{table}

\renewcommand{\arraystretch}{0.9}
\begin{table}[ht]
\caption{Average $R^2$ score divided by highest average $R^2$ score per dataset (III/III)}
\label{tab:empirical_results_p3}
\centering
\footnotesize
\begin{tabularx}{\textwidth}{llrrrrrrrr}
\toprule
source &   id &          CART &         PILOT &            RF &            RaFFLE &  \makecell{RaFFLE\\Default} &           XGB &         Ridge &         Lasso \\
\midrule
  PMLB &  657 &          0.71 &          0.92 &          0.95 &  \textbf{1.0} & \textbf{0.99} &          0.94 &          0.23 &          0.25 \\
  PMLB &  658 &          0.51 &          0.88 &          0.81 &  \textbf{1.0} & \textbf{0.99} &          0.84 &          0.13 &          0.18 \\
  PMLB &  659 &          0.11 &          0.92 &          0.94 &  \textbf{1.0} & \textbf{0.98} &          0.86 &          0.94 &          0.94 \\
  PMLB &  663 & \textbf{0.98} & \textbf{0.99} & \textbf{0.99} &  \textbf{1.0} &  \textbf{1.0} &  \textbf{1.0} & \textbf{0.97} & \textbf{0.97} \\
  PMLB &  665 &           0.0 &          0.79 &          0.82 &  \textbf{1.0} &          0.95 &           0.0 &          0.87 &          0.89 \\
  PMLB &  666 &          0.21 &          0.94 &          0.94 &  \textbf{1.0} & \textbf{0.99} &          0.81 &          0.88 &          0.88 \\
  PMLB &  678 &           0.0 &          0.91 &          0.19 &          0.87 &          0.52 &          0.22 &  \textbf{1.0} &  \textbf{1.0} \\
  PMLB &  687 &          0.22 &          0.87 & \textbf{0.96} &  \textbf{1.0} &          0.88 &          0.79 &           0.8 &          0.85 \\
  PMLB &  690 & \textbf{0.98} &  \textbf{1.0} &  \textbf{1.0} &  \textbf{1.0} &  \textbf{1.0} &  \textbf{1.0} &          0.92 &          0.92 \\
  PMLB &  695 &          0.86 &          0.94 &          0.95 & \textbf{0.99} & \textbf{0.96} &          0.92 &  \textbf{1.0} &  \textbf{1.0} \\
  PMLB &  706 &          0.25 &          0.95 &          0.92 &          0.92 &          0.81 &          0.63 &  \textbf{1.0} & \textbf{0.99} \\
  PMLB &  712 &           0.7 & \textbf{0.98} & \textbf{0.96} &  \textbf{1.0} & \textbf{0.99} &          0.79 &  \textbf{1.0} &  \textbf{1.0} \\
  PMLB & 1027 &          0.89 & \textbf{0.99} & \textbf{0.98} &  \textbf{1.0} &  \textbf{1.0} &          0.94 &  \textbf{1.0} &  \textbf{1.0} \\
  PMLB & 1028 &          0.89 &          0.75 & \textbf{0.98} & \textbf{0.99} & \textbf{0.98} &          0.89 &  \textbf{1.0} &  \textbf{1.0} \\
  PMLB & 1029 &          0.91 &  \textbf{1.0} &          0.95 &  \textbf{1.0} &  \textbf{1.0} &          0.91 &  \textbf{1.0} &  \textbf{1.0} \\
  PMLB & 1030 &          0.92 & \textbf{0.98} &          0.94 &  \textbf{1.0} &  \textbf{1.0} &          0.92 & \textbf{0.99} & \textbf{0.99} \\
  PMLB & 1089 &          0.76 & \textbf{0.98} &          0.92 &  \textbf{1.0} & \textbf{0.97} &          0.89 &          0.93 &          0.94 \\
  PMLB & 1096 &          0.65 & \textbf{0.96} &          0.92 & \textbf{0.99} & \textbf{0.98} &          0.94 & \textbf{0.96} &  \textbf{1.0} \\
  PMLB & 1193 &          0.35 & \textbf{0.98} & \textbf{0.99} &  \textbf{1.0} &  \textbf{1.0} & \textbf{0.98} &          0.95 &          0.95 \\
  PMLB & 1199 &           0.0 &          0.93 &  \textbf{1.0} & \textbf{0.99} & \textbf{0.98} &          0.93 &          0.93 &          0.93 \\
  PMLB & 1201 &           0.0 &          0.81 &          0.75 & \textbf{0.97} & \textbf{0.97} &  \textbf{1.0} &          0.37 &          0.37 \\
  PMLB & 1203 &          0.77 & \textbf{0.99} & \textbf{0.99} &  \textbf{1.0} &  \textbf{1.0} &  \textbf{1.0} &          0.85 &          0.85 \\
  PMLB & 4544 &          0.43 &           0.9 &          0.91 & \textbf{0.98} &          0.93 &          0.87 & \textbf{0.99} &  \textbf{1.0} \\
   UCI &    1 &          0.15 &          0.93 & \textbf{0.98} &  \textbf{1.0} & \textbf{0.99} &          0.88 &          0.95 &          0.95 \\
   UCI &    9 &          0.88 &          0.95 & \textbf{0.99} &  \textbf{1.0} &  \textbf{1.0} & \textbf{0.98} &          0.92 &          0.86 \\
   UCI &   10 &          0.71 &          0.58 &  \textbf{1.0} &          0.95 &          0.92 &           0.9 &          0.72 &          0.34 \\
   UCI &   60 &           0.0 &          0.36 &          0.94 &  \textbf{1.0} &          0.92 &           0.0 &          0.42 &          0.51 \\
   UCI &   87 &          0.92 &          0.72 & \textbf{0.98} &          0.94 &          0.93 &  \textbf{1.0} &          0.79 &          0.79 \\
   UCI &  165 &          0.91 &           0.9 & \textbf{0.98} & \textbf{0.99} & \textbf{0.98} &  \textbf{1.0} &          0.65 &          0.65 \\
   UCI &  183 &          0.46 &          0.93 & \textbf{0.97} &  \textbf{1.0} & \textbf{0.99} &          0.91 & \textbf{0.98} & \textbf{0.98} \\
   UCI &  186 &          0.14 &          0.61 & \textbf{0.98} &  \textbf{1.0} &          0.95 &          0.89 &          0.54 &          0.53 \\
   UCI &  275 &          0.91 &          0.95 & \textbf{0.98} &  \textbf{1.0} &  \textbf{1.0} & \textbf{0.99} &          0.73 &          0.73 \\
   UCI &  291 &           0.9 &          0.84 & \textbf{0.98} &  \textbf{1.0} & \textbf{0.96} &  \textbf{1.0} &          0.48 &          0.16 \\
   UCI &  294 & \textbf{0.96} & \textbf{0.98} &  \textbf{1.0} &  \textbf{1.0} &  \textbf{1.0} &  \textbf{1.0} & \textbf{0.96} & \textbf{0.96} \\
   UCI &  332 &           0.0 &          0.79 &  \textbf{1.0} &          0.92 &           0.0 &          0.77 &          0.78 &          0.52 \\
   UCI &  368 &          0.88 &          0.92 &          0.89 &          0.92 &          0.92 &           0.9 &  \textbf{1.0} & \textbf{0.97} \\
   UCI &  374 &          0.25 &          0.33 &  \textbf{1.0} & \textbf{0.97} &          0.94 & \textbf{0.97} &           0.3 &           0.3 \\
   UCI &  381 &           0.8 &          0.78 & \textbf{0.99} &  \textbf{1.0} & \textbf{0.98} & \textbf{0.99} &          0.32 &          0.32 \\
   UCI &  409 &          0.68 &          0.92 &          0.87 & \textbf{0.98} &          0.94 &          0.92 &  \textbf{1.0} &          0.83 \\
   UCI &  477 &          0.66 &          0.91 &  \textbf{1.0} & \textbf{0.99} & \textbf{0.98} & \textbf{0.97} &          0.76 &          0.67 \\
   UCI &  492 &           0.0 &          0.63 &          0.82 & \textbf{0.96} &          0.82 &  \textbf{1.0} &           0.0 &          0.23 \\
   UCI &  597 &          0.81 &          0.93 & \textbf{0.99} & \textbf{0.99} & \textbf{0.99} &  \textbf{1.0} &          0.93 &          0.77 \\
\midrule
       & average &     0.64 &          0.91 &          0.92 & \textbf{0.99} & \textbf{0.96} &           0.9 &          0.59 &          0.58 \\
       &  std &          0.28 &          0.15 &          0.09 &          0.02 &           0.1 &          0.15 &          0.33 &          0.31 \\
\bottomrule
\end{tabularx}
\end{table}


\clearpage

\noindent{\bf Software availability.} The \texttt{Python} and \texttt{C++} code for RaFFLE, and an example script, are at
\url{ https://github.com/STAN-UAntwerp/PILOT/tree/raffle-paper-clean}.


\begin{thebibliography}{}

\bibitem[\protect\citeauthoryear{Ao, Li, Zhu, Ali, and Yang}{Ao et~al.}{2019}]{ao2019linear}
Ao, Y., H.~Li, L.~Zhu, S.~Ali, and Z.~Yang (2019).
\newblock The linear random forest algorithm and its advantages in machine learning assisted logging regression modeling.
\newblock {\em Journal of Petroleum Science and Engineering\/}~{\em 174}, 776--789.

\bibitem[\protect\citeauthoryear{Arik and Pfister}{Arik and Pfister}{2021}]{tabnet}
Arik, S.~{\"O}. and T.~Pfister (2021).
\newblock Tabnet: Attentive interpretable tabular learning.
\newblock In {\em Proceedings of the AAAI conference on artificial intelligence}, Volume~35, pp.\  6679--6687.

\bibitem[\protect\citeauthoryear{Breiman}{Breiman}{2001}]{breiman2001random}
Breiman, L. (2001).
\newblock Random forests.
\newblock {\em Machine Learning\/}~{\em 45\/}(1), 5--32.

\bibitem[\protect\citeauthoryear{Breiman, Friedman, Olshen, and Stone}{Breiman et~al.}{1984}]{cart}
Breiman, L., J.~H. Friedman, R.~A. Olshen, and C.~J. Stone (1984).
\newblock {\em Classification and regression trees}.
\newblock Milton Park: Routledge.

\bibitem[\protect\citeauthoryear{Chen and Guestrin}{Chen and Guestrin}{2016}]{xgboost}
Chen, T. and C.~Guestrin (2016).
\newblock Xgboost: A scalable tree boosting system.
\newblock In {\em Proceedings of the 22nd ACM SIGKDD International Conference on Knowledge Discovery and Data Mining}.

\bibitem[\protect\citeauthoryear{Cortez and Morais}{Cortez and Morais}{2007a}]{Cortez2007ADM}
Cortez, P. and A.~Morais (2007a).
\newblock A data mining approach to predict forest fires using meteorological data.
\newblock In {\em Proceedings of the 13th Portuguese Conference on Artificial Intelligence}, pp.\  512--523.

\bibitem[\protect\citeauthoryear{Cortez and Morais}{Cortez and Morais}{2007b}]{forest_fires_162}
Cortez, P. and A.~Morais (2007b).
\newblock {Forest Fires}.
\newblock UCI Machine Learning Repository.
\newblock {DOI}: https://doi.org/10.24432/C5D88D.

\bibitem[\protect\citeauthoryear{Fernandes, Vinagre, Cortez, and Sernadela}{Fernandes et~al.}{2015}]{online_news_popularity_332}
Fernandes, K., P.~Vinagre, P.~Cortez, and P.~Sernadela (2015).
\newblock {Online News Popularity}.
\newblock UCI Machine Learning Repository.
\newblock {DOI}: https://doi.org/10.24432/C5NS3V.

\bibitem[\protect\citeauthoryear{Fernández-Delgado, Sirsat, Cernadas, Alawadi, Barro, and Febrero-Bande}{Fernández-Delgado et~al.}{2019}]{FERNANDEZDELGADO201911}
Fernández-Delgado, M., M.~Sirsat, E.~Cernadas, S.~Alawadi, S.~Barro, and M.~Febrero-Bande (2019).
\newblock An extensive experimental survey of regression methods.
\newblock {\em Neural Networks\/}~{\em 111}, 11--34.

\bibitem[\protect\citeauthoryear{Freund and Schapire}{Freund and Schapire}{1996}]{freund1996experiments}
Freund, Y. and R.~E. Schapire (1996).
\newblock Experiments with a new boosting algorithm.
\newblock In {\em Proceedings of the Thirteenth International Conference on Machine Learning}, pp.\  148--156. ACM.

\bibitem[\protect\citeauthoryear{Gy{\"o}rfi, Kohler, Krzyzak, and Walk}{Gy{\"o}rfi et~al.}{2002}]{distribution}
Gy{\"o}rfi, L., M.~Kohler, A.~Krzyzak, and H.~Walk (2002).
\newblock {\em A Distribution-{F}ree Theory of Nonparametric Regression}.
\newblock New York: Springer.

\bibitem[\protect\citeauthoryear{Ke, Meng, Finley, Wang, Chen, Ma, Ye, and Liu}{Ke et~al.}{2017}]{lightgbm}
Ke, G., Q.~Meng, T.~Finley, T.~Wang, W.~Chen, W.~Ma, Q.~Ye, and T.-Y. Liu (2017).
\newblock {LightGBM: A Highly Efficient Gradient Boosting Decision Tree}.
\newblock In {\em Neural Information Processing Systems}.

\bibitem[\protect\citeauthoryear{Kelly, Longjohn, and Nottingham}{Kelly et~al.}{2024}]{uciml}
Kelly, M., R.~Longjohn, and K.~Nottingham (2024).
\newblock Uci machine learning repository.

\bibitem[\protect\citeauthoryear{Klusowski and Tian}{Klusowski and Tian}{2024}]{klusowski2024large}
Klusowski, J.~M. and P.~M. Tian (2024).
\newblock Large scale prediction with decision trees.
\newblock {\em Journal of the American Statistical Association\/}~{\em 119\/}(545), 525--537.

\bibitem[\protect\citeauthoryear{Li and Li}{Li and Li}{2011}]{li2011learning}
Li, C. and H.~Li (2011).
\newblock Learning random model trees for regression.
\newblock {\em International Journal of Computers and Applications\/}~{\em 33\/}(3), 258--265.

\bibitem[\protect\citeauthoryear{Loh}{Loh}{2011}]{loh2011classification}
Loh, W.-Y. (2011).
\newblock Classification and regression trees.
\newblock {\em Wiley Interdisciplinary Reviews: Data Mining and Knowledge Discovery\/}~{\em 1\/}(1), 14--23.

\bibitem[\protect\citeauthoryear{Mendes-Moreira, Soares, Jorge, and Sousa}{Mendes-Moreira et~al.}{2012}]{mendes2012ensemble}
Mendes-Moreira, J., C.~Soares, A.~M. Jorge, and J.~F.~D. Sousa (2012).
\newblock Ensemble approaches for regression: A survey.
\newblock {\em ACM Computing Surveys\/}~{\em 45\/}(1), 1--40.

\bibitem[\protect\citeauthoryear{Olson, La~Cava, Orzechowski, Urbanowicz, and Moore}{Olson et~al.}{2017}]{Olson2017PMLB}
Olson, R.~S., W.~La~Cava, P.~Orzechowski, R.~J. Urbanowicz, and J.~H. Moore (2017, Dec).
\newblock {PMLB: A large benchmark suite for machine learning evaluation and comparison}.
\newblock {\em BioData Mining\/}~{\em 10\/}(36), 1--13.

\bibitem[\protect\citeauthoryear{Pedregosa, Varoquaux, Gramfort, Michel, Thirion, Grisel, Blondel, Prettenhofer, Weiss, Dubourg, Vanderplas, Passos, Cournapeau, Brucher, Perrot, and Duchesnay}{Pedregosa et~al.}{2011}]{scikit-learn}
Pedregosa, F., G.~Varoquaux, A.~Gramfort, V.~Michel, B.~Thirion, O.~Grisel, M.~Blondel, P.~Prettenhofer, R.~Weiss, V.~Dubourg, J.~Vanderplas, A.~Passos, D.~Cournapeau, M.~Brucher, M.~Perrot, and E.~Duchesnay (2011).
\newblock Scikit-learn: Machine learning in {P}ython.
\newblock {\em Journal of Machine Learning Research\/}~{\em 12}, 2825--2830.

\bibitem[\protect\citeauthoryear{Raymaekers, Rousseeuw, Verdonck, and Yao}{Raymaekers et~al.}{2024}]{raymaekers2024pilot}
Raymaekers, J., P.~J. Rousseeuw, T.~Verdonck, and R.~Yao (2024).
\newblock Fast linear model trees by {PILOT}.
\newblock {\em Machine Learning\/}~{\em 113}, 6561--6610.

\bibitem[\protect\citeauthoryear{Rodr{\'\i}guez, Garc{\'\i}a-Osorio, Maudes, and D{\'\i}ez-Pastor}{Rodr{\'\i}guez et~al.}{2010}]{rodriguez2010experimental}
Rodr{\'\i}guez, J.~J., C.~Garc{\'\i}a-Osorio, J.~Maudes, and J.~F. D{\'\i}ez-Pastor (2010).
\newblock An experimental study on ensembles of functional trees.
\newblock In {\em Multiple Classifier Systems: 9th International Workshop, MCS 2010, Cairo, Egypt. Proceedings 9}, pp.\  64--73. Springer.

\bibitem[\protect\citeauthoryear{Rodriguez-Galiano, Sanchez-Castillo, Chica-Olmo, and Chica-Rivas}{Rodriguez-Galiano et~al.}{2015}]{rf_performance_rodriguez2015}
Rodriguez-Galiano, V., M.~Sanchez-Castillo, M.~Chica-Olmo, and M.~Chica-Rivas (2015).
\newblock Machine learning predictive models for mineral prospectivity: An evaluation of neural networks, random forests, regression trees and support vector machines.
\newblock {\em Ore Geology Reviews\/}~{\em 71}, 804--818.

\bibitem[\protect\citeauthoryear{Romano, Le, La~Cava, Gregg, Goldberg, Chakraborty, Ray, Himmelstein, Fu, and Moore}{Romano et~al.}{2022}]{romano2021pmlb}
Romano, J.~D., T.~T. Le, W.~La~Cava, J.~T. Gregg, D.~J. Goldberg, P.~Chakraborty, N.~L. Ray, D.~Himmelstein, W.~Fu, and J.~H. Moore (2022).
\newblock {PMLB v1.0: An open source dataset collection for benchmarking machine learning methods}.
\newblock {\em Bioinformatics\/}~{\em 38\/}(3), 878–880.

\bibitem[\protect\citeauthoryear{Shi, Li, and Li}{Shi et~al.}{2019}]{shi2019gradient}
Shi, Y., J.~Li, and Z.~Li (2019).
\newblock Gradient boosting with piece-wise linear regression trees.
\newblock In {\em Proceedings of the 28th International Joint Conference on Artificial Intelligence}, pp.\  3432--3438.

\bibitem[\protect\citeauthoryear{Song, Langfelder, and Horvath}{Song et~al.}{2013}]{song2013random}
Song, L., P.~Langfelder, and S.~Horvath (2013).
\newblock Random generalized linear model: {A} highly accurate and interpretable ensemble predictor.
\newblock {\em BMC Bioinformatics\/}~{\em 14}, 1--22.

\bibitem[\protect\citeauthoryear{Stulp and Sigaud}{Stulp and Sigaud}{2015}]{stulp2015many}
Stulp, F. and O.~Sigaud (2015).
\newblock Many regression algorithms, one unified model: A review.
\newblock {\em Neural Networks\/}~{\em 69}, 60--79.

\bibitem[\protect\citeauthoryear{Torgo}{Torgo}{1997}]{torgo1997functional}
Torgo, L. (1997).
\newblock Functional models for regression tree leaves.
\newblock In {\em Proceedings of the Fourteenth International Conference on Machine Learning}, pp.\  385--393.

\bibitem[\protect\citeauthoryear{Zamo, Mestre, Arbogast, and Pannekoucke}{Zamo et~al.}{2014}]{zamo2014benchmark}
Zamo, M., O.~Mestre, P.~Arbogast, and O.~Pannekoucke (2014).
\newblock A benchmark of statistical regression methods for short-term forecasting of photovoltaic electricity production, {P}art {I}: {D}eterministic forecast of hourly production.
\newblock {\em Solar Energy\/}~{\em 105}, 792--803.

\end{thebibliography}

\end{document}